\documentclass[conference]{IEEEtran}
\IEEEoverridecommandlockouts
\usepackage{cite}
\usepackage{amsmath,amssymb,amsfonts}
\usepackage{graphicx}
\usepackage{textcomp}
\usepackage{xcolor}

\usepackage{booktabs}
\usepackage{multirow}
\usepackage{amsmath, amssymb}
\usepackage{algorithm, algpseudocode}
\usepackage{enumitem}
\usepackage{hyperref}
\usepackage{cleveref}
\usepackage{nicefrac}
\usepackage{tikz}
\usetikzlibrary{arrows.meta, positioning, shapes.geometric, fit,
                backgrounds, calc}
\usepackage{pgfplots}
\pgfplotsset{compat=1.18}
\usepgfplotslibrary{groupplots}

\hypersetup{
    colorlinks=true,
    linkcolor=blue!50!black,
    citecolor=blue!50!black,
    urlcolor=blue!50!black,
}

\def\BibTeX{{\rm B\kern-.05em{\sc i\kern-.025em b}\kern-.08em
    T\kern-.1667em\lower.7ex\hbox{E}\kern-.125emX}}
\begin{document}

\title{Privacy Policy Enforcement Guardrails for Data-Sensitive Retrieval-Augmented Generation
}

\author{
\IEEEauthorblockN{Osama Zafar}
\IEEEauthorblockA{\textit{Dept. of Computer and Data Sciences} \\
\textit{Case Western Reserve University}\\
Cleveland, USA \\
oxz23@case.edu}
\and
\IEEEauthorblockN{Alexander Nemecek}
\IEEEauthorblockA{\textit{Dept. of Computer and Data Sciences} \\
\textit{Case Western Reserve University}\\
Cleveland, USA \\
ajn98@case.edu}
\and
\IEEEauthorblockN{Yiqian Zhang}
\IEEEauthorblockA{\textit{Dept. of Computer and Data Sciences} \\
\textit{Case Western Reserve University}\\
Cleveland, USA \\
yxz3283@case.edu}
\and
\IEEEauthorblockN{Wenbiao Li}
\IEEEauthorblockA{\textit{Dept. of Computer and Data Sciences} \\
\textit{Case Western Reserve University}\\
Cleveland, USA \\
wxl387@case.edu}
\and
\IEEEauthorblockN{Debargha Ganguly}
\IEEEauthorblockA{\textit{Dept. of Computer and Data Sciences} \\
\textit{Case Western Reserve University}\\
Cleveland, USA \\
dxg512@case.edu}
\and
\IEEEauthorblockN{Vikash Singh}
\IEEEauthorblockA{\textit{Dept. of Computer and Data Sciences} \\
\textit{Case Western Reserve University}\\
Cleveland, USA \\
vxs465@case.edu}
\and
\IEEEauthorblockN{Vipin Chaudhary}
\IEEEauthorblockA{\textit{Dept. of Computer and Data Sciences} \\
\textit{Case Western Reserve University}\\
Cleveland, USA \\
vxc204@case.edu}
\and
\IEEEauthorblockN{Erman Ayday}
\IEEEauthorblockA{\textit{Dept. of Computer and Data Sciences} \\
\textit{Case Western Reserve University}\\
Cleveland, USA \\
exa208@case.edu}
}

\maketitle

\begin{abstract}
Retrieval-augmented generation (RAG) has become the default deployment pattern in medicine, finance, and law, domains where every generated answer is grounded in corpora that are sensitive in nature. The privacy defenses are primarily constructed under the assumption that data leakage manifests as a Social Security number, an email address, or a name. However, this is not invariably the case. The most perilous form of leakage is contextual in nature, which standard token-level Personally Identifiable Information (PII) filters fail to detect. For instance, the sentence “52-year-old female litigator from Omaha with two children and a recent custody filing” contains five attributes, none of which are individually regulated, that collectively identify a single individual. Token-level filters and named-entity pipelines are blind to this by construction, and emerging representation-based detectors often learn the style of leaky text rather than the leak itself, achieving high benchmark scores while missing the underlying threat.

We address this gap with a privacy policy enforcement framework (PPE) for quasi-identifier (QI) cluster leakage in RAG outputs, built on a dual one-class density estimator over fused text embeddings with a calibrated abstain region for inputs that are out-of-distribution to both the safe and unsafe densities. Because real breaches cannot provide training data at the required scale, we introduce an axis-stratified (non-overlapping subsets of a domain), multi-LLM synthetic data generation pipeline and instantiate it across three independent domains: medicine, finance, and law. This three-domain setup exposes a case-style confound: a Gaussian Mixture baseline reaches AUROC $\geq$ 0.94 within-distribution, but collapses to [0.72 - 0.79] on a borderline-safe stress test that strips the QIs but maintains the tone referring to the client, patient, or matter. The detector has been latching onto the register, not content. A T3 detector (proposed by Ganguly et al. \cite{ganguly2026t3}) with one-class SVM (T3+OCSVM) trained on the safe distribution data, augmented with borderline-safe records, achieves a promising borderline AUROC of 0.93 or higher across all three domains. It also significantly lowers the borderline false positive rate at a 95\% true positive rate (FPR$_{95}$) by 44–55 percentage points, while keeping abstention rates below 11\% and operating at millisecond-level latency. Our framework is benchmarked against a supervised MLP classifier and a 14-billion-parameter large language model serving as an evaluative judge. While they offer comparable levels of accuracy, neither is appropriate for operational deployment. The supervised classifier abstains in 70–85\% of borderline sensitive cases, and the LLM judge cannot be calibrated to a specific operational point and exhibits significant latency. The borderline-safe stress test we propose as a methodological contribution extends beyond contextual privacy and applies to any classifier trained on synthetically generated positive and negative classes.
\end{abstract}


\section{Introduction}

Retrieval-augmented generation (RAG) changes the unit at which privacy must be enforced. In a conventional system, privacy policies attach to stored objects: a patient note, a transaction record, a legal memorandum. In a RAG system, the object exposed to the user is a generated answer. That answer may quote, summarize, combine, or infer from protected records, and its privacy status is not automatically inherited from the access controls on the underlying corpus. The \emph{privacy enforcement} problem, therefore, moves from document access to output composition. 

The easy case to handle is direct leakage, the explicit disclosure of a name, Social Security number, medical record number, account number, docket number, e-mail address, or phone number. Existing personally identifiable information PII filters, NER systems, and regular-expression pipelines provide a natural first line of defense (Layer 1). These mechanisms are imperfect, but the violation has a local signature: some token or span is itself disallowed. The harder case is contextual leakage. Consider an answer that describes a ``52-year-old female litigator from Omaha with two children and a recent custody filing.'' No token is a direct identifier. Yet the conjunction may be identifying \cite{sweeney2002kanonymity, carlini2021extracting}. The privacy violation is not a string; it is a composition of quasi-identifiers. This constitutes the Layer-2 challenge explored here: detection of re-identifying QI clusters in generated RAG outputs that have already passed direct-identifier filtering.

At first glance, contextual privacy appears to be a natural setting for distribution-based guardrails like \emph{Trust the Typical} (T3)~\cite{ganguly2026t3}. T3 builds guardrails by learning what acceptable behavior typically looks like and rejecting outputs outside that support. This is attractive for RAG privacy because real privacy breaches are rare, sensitive, and unavailable at the training scale. But T3 is calibrated against the assumption that unsafe content looks atypical, and that assumption fails here. The difficulty is that privacy-violating answers do not necessarily look unusual. In clinical, financial, and legal RAG, case-oriented professional language is not an anomaly. A leaky answer may therefore be typical in tone, domain, and task. What makes it unsafe is not its style, but the way it binds ordinary attributes; age, location, occupation, family structure, assets, diagnoses, or matter details to a referential subject. The guardrail has to distinguish between professional text that is non-identifying and professional text whose attribute composition points back to a person, household, client, patient, or matter. Applied naively, a T3-style detector trained against synthetic unsafe examples in patient/client/matter voice and synthetic safe examples in public educational voice will separate \emph{registers}, not privacy boundaries, while reporting high within-distribution AUROC. Another challenge is the data itself, real-world RAG breach data does not exist at the scale required to train such a detector, so the unsafe class must be synthesized.

We address this with a layered framework. First, we construct \emph{borderline-safe} examples: counterfactual negatives that preserve the professional voice of the unsafe class but remove the QI clusters. They turn the synthetic benchmark into a stress test that asks whether the detector responds to composition once style is no longer predictive of the label. Second, we situate the stress test inside an axis-stratified (non-overlapping subsets of a domain), multi-LLM pipeline that produces aligned safe, unsafe, and borderline-safe partitions across medicine, finance, and law, so the diagnostic and the fix can be evaluated against domain variation. Third, we extend T3's safe-side typicality with a parallel one-class detector on the unsafe side, exposing a calibrated abstention region for inputs that lie in neither distribution, a regime that T3-style guardrails alone do not characterize. Finally, we benchmark this dual-density detector against a supervised MPL classifier and a 14B LLM-as-judge on the same evaluation substrate.

This paper makes three main contributions.

\begin{enumerate}
    \item[\textbf{C1}]~\textbf{An axis-stratified synthetic data generation pipeline}. We introduce a multi-LLM generation framework that transforms public seed corpora into aligned safe, unsafe, and borderline-safe Q\&A partitions. It is parameterized by a seed corpus, an 8-class Quasi-Identifier (QI) taxonomy, and domain-specific validators. We instantiate it across three high-stakes domains: medicine, finance, and law (see Section~\ref{sec:data}, Tables~\ref{tab:seedcorpus} and~\ref{tab:gen}).
    \item[\textbf{C2}] \textbf{The case-style confound and the borderline-safe stress test.} We provide a critical diagnostic finding: across three independently constructed domains, our Gaussian Mixture baseline achieves within-distribution AUROC $geq$ 0.94; however, it collapses to [0.72, 0.79] on a borderline-safe stress test that strips quasi-identifiers but preserves the referential tone, indicating that the T3+GMM detector had been measuring style rather than content. We propose the borderline-safe stress test as a methodological contribution that extends beyond contextual privacy to any classifier trained on synthetically constructed positive and negative classes (see Section~\ref{sec:eval:borderline} and Figure~\ref{fig:results}).
    \item[\textbf{C3}] \textbf{A Robust Solution.} We present a novel solution to mitigate stylistic confounding: a one-class SVM estimator on the safe side, augmented with 4,000 borderline-safe records during training. This recovers a borderline AUROC to $\geq 0.93$ across all three domains, reduces borderline FPR95 by 44–55 percentage points, and keeps abstain rates below 11\% at the operational threshold, below which Layer 2 becomes deployable rather than aspirational. (see Section~\ref{sec:disc:generalization}).
\end{enumerate}

The remainder of this paper is organized as follows. We begin in Section~\ref{sec:related} by surveying the background and relevant literature, followed by a formalization of the threat model in Section~\ref{sec:threat}. A high-level system overview is provided in Section~\ref{sec:overview} (Figure~\ref{fig:system}). Detailed specifications for our axis-stratified data pipeline (Figure~\ref{fig:datapipeline}) and the dual-density detector architecture (Figure~\ref{fig:detector}) are presented in Section~\ref{sec:data} and Section~\ref{sec:detection}, respectively. Our experimental setup and core results are reported in Section~\ref{sec:eval}. Section~\ref{sec:comparison} compares the proposed detector to two architectural alternatives (a supervised binary classifier and an LLM-as-judge baseline). Section~\ref{sec:disc} synthesizes the cross-domain and cross-architecture findings and discusses operating-point selection. We address limitations in Section~\ref{sec:limitations} and offer concluding remarks in Section~\ref{sec:conclusion}. Supplementary per-domain taxonomies and implementation details are provided in Appendix~\ref{app:domains}.

\section{Background and Related Work}
\label{sec:related}
The integration of Large Language Models (LLMs) into clinical and professional workflows has necessitated robust frameworks for privacy preservation. This section situates our work within the evolving landscape of RAG security, safety guardrails, and statistical anomaly detection.

\subsection{Privacy Risks and Memorization in LLMs}

A foundational body of research has demonstrated that LLMs are prone to memorizing sensitive training data which can subsequently be extracted through various adversarial techniques \cite{carlini2021extracting, carlini2023quantifying}. Extraction attacks have revealed that even infrequent strings in the training set can be elicited via prefix-based prompting \cite{nasr2023scalable}. Membership inference attacks (MIA) further allow an adversary to determine if a specific record was present in the model history \cite{shokri2017membership, mattern2023membership}, including domain-specific instantiations against clinical LLMs fine-tuned on EHR data \cite{nemecek2025exploring}. PPE addresses a distinct threat model where leakage occurs during inference from the external knowledge base retrieved in RAG settings.

\subsection{Security in Retrieval-Augmented Generation}

RAG shifts the privacy focus from training-time memorization to inference-time disclosure \cite{lewis2020rag, guu2020realm}. Dynamic repositories introduce real-time elution risks where the retriever serves as a conduit for sensitive information. Recent studies have demonstrated the feasibility of extracting retrieved documents through prompt injection and characterized the unique privacy issues inherent in RAG datastores \cite{zeng2024good, cohen2024compromptmized}. Furthermore, membership inference attacks and tool-use vulnerabilities have been successfully adapted to target retrieved chunks, expanding the observational surface across heterogeneous repositories \cite{yao2023react, schick2023toolformer, li2024mia}. Orthogonal cryptographic approaches verify training-data provenance through zero-knowledge proofs, ensuring models were fine-tuned on authorized datasets \cite{namazi2025zkprov}. Whereas prior work has focused on securing the retrieval pipeline or attesting to its inputs, PPE targets the generated output as the final point of policy enforcement

\subsection{The Limits of Classical De-identification}

Privacy preservation in clinical data has traditionally relied on de-identification standards such as k-anonymity, which posits that an individual is protected if their attributes are indistinguishable from others in a cohort \cite{hipaa1996, sweeney2002kanonymity}. Refinements such as l-diversity and t-closeness addressed vulnerabilities related to attribute skewness \cite{machanavajjhala2007ldiversity, li2007tcloseness}. However, these classical methods often fail for unstructured text, where aggregate narratives facilitate re-identification despite token-level redaction \cite{elemam2011systematic, rocher2019estimating}. Although PII benchmarks like the Text Anonymization Benchmark (TAB) have been developed to measure de-identification effectiveness, they often lack the scale or domain diversity required for multi-agent RAG systems \cite{lukas2023analyzing}. PPE fills this gap by utilizing a large-scale synthetic pipeline to generate domain-specific quasi-identifier clusters.

\subsection{Reactive Guardrails and Safety Filters}

To govern model behavior, developers have introduced diverse alignment techniques and system-level guardrails. Methods such as RLHF and Constitutional AI attempt to embed safety norms into model weights \cite{ouyang2022training, bai2022constitutional}. External guardrails like Llama Guard and NeMo Guardrails provide a modular defense layer by classifying interactions against predefined safety taxonomies \cite{inan2023llama, rebedea2023nemo}. Notably, these guardrails are trained on harm taxonomies that typically do not include contextual privacy violations, creating a structural gap in their enforcement logic \cite{han2024wildguard, zou2023universal}. PPE positions contextual privacy as a first-class safety concern, closing the gap left by general-purpose moderation tools \cite{wei2023jailbroken}.

\subsection{Density Estimation and Statistical Typicality}

Identifying subtle semantic anomalies is a core task in out-of-distribution (OOD) detection, a form of anomaly detection. Traditional methods range from parametric Gaussian Mixture Models (GMM) to non-parametric approaches like one-class SVM and Deep SVDD \cite{mclachlan1988mixture, scholkopf1999support, ruff2018deep}. In generative AI, information-theoretic principles have introduced statistical typicality as a robust metric for uncertainty quantification in computer vision\cite{cover2006elements, nalisnick2019detecting, morningstar2021density, gangulyForteFindingOutliers2024a}. Typicality-based methods evaluate whether a sample resides within the concentrated typical set of the target distribution, and are used for MRI images \cite{gangulyForteFindingOutliers2024a}, HPC log anomaly detection \cite{chen2025k}, quality control in automated agent workflows\cite{ganguly2025labeling}, and material science workflows\cite{lu2026context}. Recent work on the T3 framework has operationalized this for language model guardrailing \cite{ganguly2026t3}. PPE extends this by applying typicality to contextual privacy and introducing a dual-density estimator to resolve stylistic confounds that T3-style prompts do not encounter (an extreme near-OOD task \cite{ren2019likelihood}).
\section{System Overview}
\label{sec:threat}

\subsection{Setting}

We consider a deployed RAG system that combines (i)~a retrieval index over a domain-specific corpus containing sensitive data (e.g., medical case notes, financial records, legal matter files); (ii)~a black-box generative LLM, with no assumption of access to fine-tuning, gradients, or internal activations; and (iii)~a privacy-enforcement mechanism (the filter) that monitors the system’s output. We further assume that the retrieval index has undergone an initial curation pass to remove direct identifiers; consequently, our focus is restricted to leakages inherent to the retrieved content itself rather than the index architecture. The filter operates exclusively on the generated answer text at inference time. It is the final gatekeeper between the model and the user, and it is the only component in the pipeline whose decisions are observable, calibratable, and replaceable without retraining the generator. This positioning is deliberate: defenses earlier in the pipeline (corpus curation, retrieval-side filtering, prompt-level constraints) are complementary, but cannot catch what the generator produces de novo through synthesis or fabrication.

\subsection{Threat Model}

Our threat model includes two adversaries, neither of which has white-box access to the filter.

\begin{itemize}
    \item The \textbf{end user} submits questions that aim to elicit privacy-relevant answers. These questions can be harmless, such as a healthcare professional asking about treatment options, or a lawyer researching procedural details, or adversarial, with the user probing for QI clusters about a specific individual. The filter is designed to be intent-agnostic, meaning it does not attempt to guess whether a query text is malicious. Instead, it evaluates every generated answer on its own privacy merits. This approach is the optimal security practice for a Layer 2 defense, as user-intent inference belongs upstream.
    \item The \textbf{generative LLM} is also treated as an unreliable generator that may produce a re-identifying answer through three primary failure modes: (i)~\textit{Retention}:~the model reproduces QI clusters that were present in the retrieved segments and bypassed the initial curation pass. (ii)~\textit{Synthesis}:~the model aggregates multiple, individually innocuous QIs from disparate retrieved chunks into a single, identifying cluster. (iii)~\textit{Fabrication}:~the model hallucinates quasi-identifiers that, despite being synthetic, appear identifying and violate the privacy policy. 
\end{itemize}

We do \emph{not} consider an adversarial \emph{filter operator} (the
filter is assumed to be honestly-deployed) and we do \emph{not}
consider gradient-guided adversarial users with white-box access to the filter's internals.

We do not consider an adversarial filter operator (the filter is honestly deployed) and we do not consider gradient-guided adversarial users with white-box access to the filter's internals or embedding pipeline. These restrictions are deliberate: they keep the threat model aligned with the deployment realities of regulated industries, where filters run inside operator infrastructure and adversaries are external query-issuers rather than internal red-teamers.

\subsection{Layered Defense}

We frame contextual privacy enforcement as a two-layer architecture (see Figure~\ref{fig:system}):

\begin{itemize}
    \item \textbf{Layer~1: Direct-Identifier Detection.} Regex and NER pipelines reject answers containing literal identifiers such as Social Security Numbers (SSNs), Medical Record Numbers (MRNs), docket numbers, email addresses, or specific named individuals. As this represents well-studied prior art, it is not the primary focus of this work.
    \item \textbf{Layer~2: Contextual-Leakage Detection.} A combination of dual learned density-based detectors to flag answers containing quasi-identifier (QI) clusters. Crucially, this mechanism identifies leaks even when they are embedded within a referential professional voice (e.g., patient, client, or matter narratives) and contain no single token that would trigger a Layer 1 violation. The design, evaluation, and optimization of this layer constitute the core contribution of this paper.
\end{itemize}

The two layers are fundamentally complementary: Layer~1 is high-precision, low-recall on direct identifiers, while Layer~2 flags indirect-QI paraphrases and aggregate-QI re-identification.

\subsection{Problem Statement}

Formally, let $X$ be the set of generated Q\&A pairs that have passed the Layer 1 filter, and let $Y \in \{\text{safe}, \text{unsafe}\}$ be a latent binary label indicating whether the pair contains a re-identifying quasi-identifier cluster. We assume access to two exemplar training corpora:
$D_s = \{x : Y(x) = \text{safe}\}$, drawn from public Q\&A seeds and
$D_u = \{x : Y(x) = \text{unsafe}\}$, synthesized via the pipeline detailed in Section~\ref{sec:data}). 

The objective for Layer 2 is to learn a mapping $f: X \to \{\text{flag},\, \text{safe},\, \text{abstain}\}$ from $D_s$ and
$D_u$ that minimizes the expected operational cost. This cost is determined by the domain-specific asymmetry between False Negatives (failing to flag a leak) and False Positives (unnecessarily flagging safe content). When the model’s predictive confidence is low, it may instead abstain, triggering a manual review at a lower fixed cost than a privacy breach would incur.

\noindent \textbf{The Case for Density-Based Detection:} while the latent label $Y$ is observable in construction sets $D_s$ and $D_u$, it remains unobservable in a live deployment. A discriminative classifier on $\{D_s, D_u\}$ would optimize the binary classification objective directly, but we observe two empirical obstacles: (i)~the synthetic $D_u$ contains stylistic correlates of $Y = \text{unsafe}$ (referential
client voice; Section~\ref{sec:eval:borderline}) that a discriminator latches onto as an easier signal than the underlying QI presence; (ii)~density-based methods admit a principled abstention region (both-OOD) that discriminative methods do not have a natural analog for. We adopt the dual one-class density approach for these reasons; we revisit the discriminative alternative in Section~\ref{sec:limitations:discriminative}.

\subsection{Overview of Proposed Solution}
\label{sec:overview}

\begin{figure*}[] 
    \centering
    \includegraphics[height=6.2cm, width=0.9\textwidth]{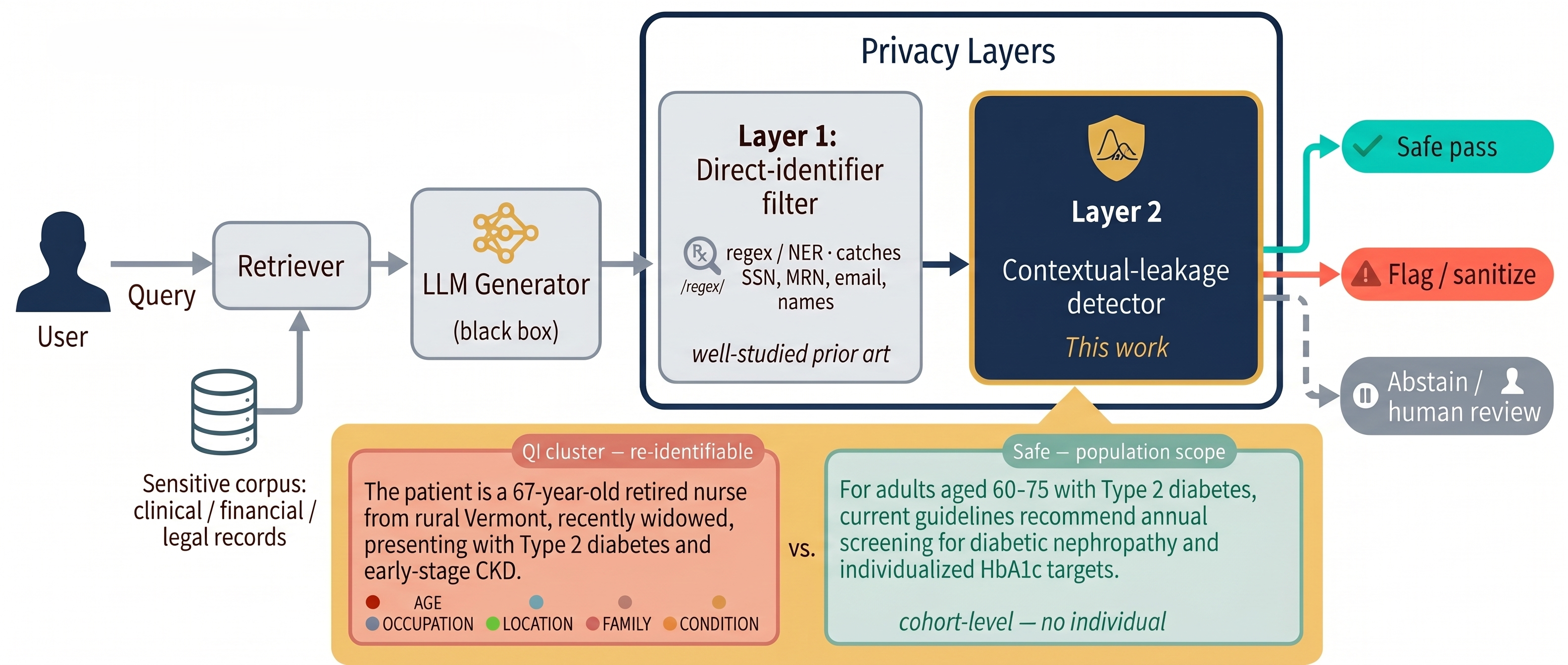} 
    \caption{Layered privacy enforcement for RAG. Layer-1 catches
    direct identifiers via regex/NER; Layer-2 (this work) detects
    contextual QI-cluster leakage and routes outputs to safe-pass,
    flag, or abstain via the discriminator $\delta$ and an explicit
    both-OOD abstain region (Section~\ref{sec:disc:operating}).}
    \label{fig:system}
\end{figure*}

We frame contextual privacy enforcement for retrieval-augmented generation (RAG) as a two-layer defense, focusing mainly on \textit{Layer 2}: the detection of quasi-identifier (QI) clusters and indirect paraphrases that bypass token-level rules yet collectively facilitate re-identification. Figure~\ref{fig:system} illustrates where Layer~2 sits in a deployed RAG pipeline. The detector processes the generated answer and emits one of three actions: \emph{flag} (sanitize or re-prompt), safe (pass through), or abstain (escalate to human review). The abstain action is triggered when an input is identified as out-of-distribution (OOD) relative to both the safe and unsafe density estimates.

\noindent \textbf{Pipeline:}
Our Layer~2 design utilizes a dual one-class density estimator over a fused text-embedding representation. Training requires only safe and unsafe \emph{exemplar} corpora for training; it avoids the need for per-record privacy labels at deployment and requires no fine-tuning of the black-box generator. Inference consists of a single forward pass through three frozen encoders followed by parallel density evaluations. The system comprises three core components:

\begin{enumerate}
    \item \textbf{Synthetic Data Pipeline} (Section~\ref{sec:data} and Figure~\ref{fig:datapipeline}): An axis-stratified framework that transforms public seed corpora into unsafe Q\&A pairs across multiple domains. Crucially, it also generates an explicit \emph{borderline-safe} corpus (genuinely-safe content that mimics each suspected stylistic confounder of the unsafe class). We instantiate this pipeline across three domains: medical, finance, and law.
    \item \textbf{Dual One-Class Detector} (Section~\ref{sec:detection} and Figure~\ref{fig:detector}): A detection engine that fuses three distinct text encoders (Qwen3, BGE-M3, and E5) and feeds the fused embedding into two density estimators (T3+T3+GMM or T3+OCSVM), one trained on safe records, one on unsafe distribution producing a discriminator $\delta = \sigma_u - \sigma_s$ thresholded against an operator-chosen $\tau$.
    \item \textbf{Borderline-Safe Stress Test} (Section~\ref{sec:eval:borderline}): A methodological framework to disentangle two competing hypotheses, whether a detector has learned the intended signal (QI presence) or a stylistic confound (e.g., the referential professional voice). The test evaluates the detector using unsafe holdouts paired with a borderline-safe corpus that is stylistically identical to the unsafe class but entirely devoid of QI clusters.
\end{enumerate}

\noindent \textbf{Notations:} Throughout the paper, $\sigma_s$ and $\sigma_u$ denote the (signed) T3+OCSVM scores against the safe and unsafe density estimates, respectively. $\delta = \sigma_u - \sigma_s$ is the discriminator score; $\tau$ is the operator threshold. \textit{v3} identifies the T3+GMM-with-cleaned-data baseline configuration; \textit{v4} identifies the T3+OCSVM-with-borderline-augmented-safe configuration (the recommended deployment configuration).

\section{Synthetic Data Generation Methodology}
\label{sec:data}

\begin{figure*}[] 
    \centering
    \includegraphics[height=6.2cm,width=0.9\textwidth]{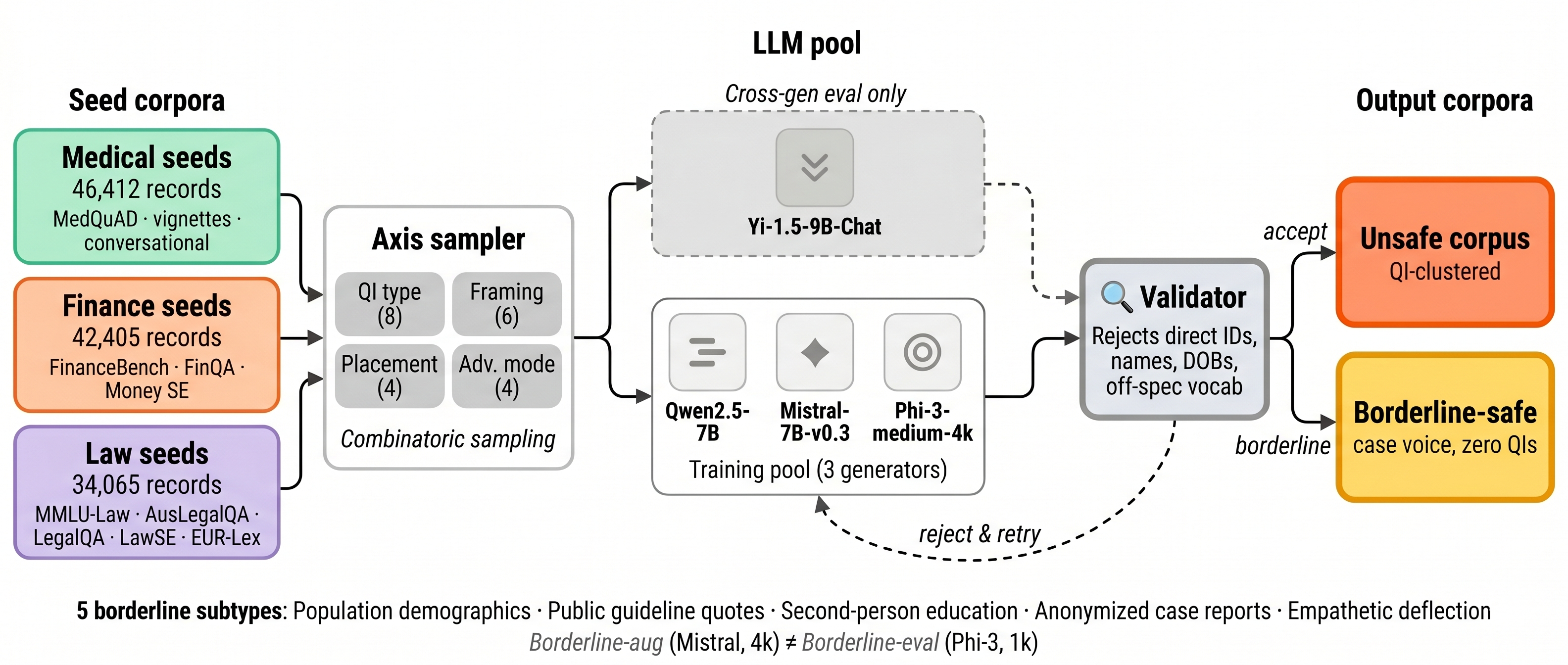} 
    \caption{Axis-stratified multi-LLM data generation pipeline,
    instantiated per domain with the QI taxonomy of
    Table~\ref{tab:qimap} and a domain-specific validator. Yi-1.5-9B
    is held out for cross-generator evaluation
    (Section~\ref{sec:eval:crossgen}); borderline-safe records are
    produced by a separate prompt template on the same generators
    (Section~\ref{sec:data:borderline}).}
    \label{fig:datapipeline}
\end{figure*}

Training a contextual-privacy detector requires paired safe and unsafe Q\&A exemplars at a scale that real-world breaches cannot provide. We utilize public seed corpora to represent the safe class and \textit{synthesize} the unsafe class through an axis-stratified, multi-LLM generation pipeline. This pipeline is instantiated across three domains i.e., medical, finance, and law, requiring only the substitution of the seed corpora, the QI taxonomy, and domain-specific validator constraints. The pipeline architecture is illustrated in Figure~\ref{fig:datapipeline}.

\subsection{Seed Corpora}
\label{sec:data:seeds}

For each domain, we select public seed corpora across three stylistic registers to mirror the distribution of a deployed RAG system: authoritative (regulator/textbook), vignette (case study), and conversational (advisor Q\&A). Table~\ref{tab:seedcorpus} summarizes these sources. Notably, all three domains are characterized by a single high-volume conversational source that dominates record counts (e.g., Stack Exchange variants). We empirically investigate how this dominance influences cross-domain residuals in Section~\ref{sec:eval:borderline}.


\begin{table}[t]
\centering
\caption{Seed-corpus composition across the three domains.}
\label{tab:seedcorpus}
\small
\setlength{\tabcolsep}{4pt}
\begin{tabular}{llrl}
\toprule
\textbf{Domain} & \textbf{Source} & \textbf{Records} & \textbf{License} \\
\midrule
\multirow{4}{*}{\textbf{Medical}}
  & MedQuAD~\cite{benabacha2019medquad}                   & 16{,}412 & CC0 1.0 \\
  & adrianf12 vignettes~\cite{adrianf12_vignettes}        & 10{,}000 & CC-BY 4.0 \\
  & kabatubare conversational~\cite{kabatubare_medical}   & 20{,}000 & CC-BY 4.0 \\
  & \emph{subtotal}                                       & \emph{46{,}412} & \\
\midrule
\multirow{4}{*}{\textbf{Finance}}
  & FinanceBench~\cite{islam2023financebench}             &     90 & CC-BY 4.0 \\
  & FinQA~\cite{chen2021finqa}                            &  8{,}280 & MIT \\
  & Money StackExchange~\cite{money_stackexchange}        & 34{,}035 & CC-BY-SA 4.0 \\
  & \emph{subtotal}                                       & \emph{42{,}405} & \\
\midrule
\multirow{6}{*}{\textbf{Law}}
  & MMLU--Law~\cite{hendrycks2021mmlu}                    &  1{,}709 & MIT \\
  & Australian Legal QA~\cite{butler2023australianlegalqa} & 2{,}124 & CC-BY 4.0 \\
  & LegalQA-v1~\cite{legalqa_v1}                          &  3{,}724 & Apache 2.0 \\
  & Law Stack Exchange~\cite{law_stackexchange}           & 25{,}366 & CC-BY-SA 4.0 \\
  & EUR-Lex-Sum (EN)~\cite{aumiller2022eurlexsum}         &  1{,}142 & CC-BY 4.0 \\
  & \emph{subtotal}                                       & \emph{34{,}065} & \\
\bottomrule
\end{tabular}
\end{table}

\subsection{Quasi-Identifier Taxonomy}
\label{sec:data:qi}

We define an 8-class QI taxonomy that preserves cardinality across domains for direct cross-domain comparative analysis. Five classes share semantics across domains (\textsc{age, occupation, location, family, affiliation}), while three classes are specialized to domain-specific identifiers that distinguish individuals from generic content (Table~\ref{tab:qimap}).


\begin{table}[t]
\centering
\caption{Eight-class quasi-identifier taxonomy. Shared classes are listed centrally; domain-specific mappings are grouped by sector.}
\label{tab:qimap}
\small
\begin{tabular}{@{}llll@{}}
\toprule
\textbf{Class} & \textbf{Medical} & \textbf{Finance} & \textbf{Law} \\ 
\midrule
\multicolumn{4}{c}{\textbf{Shared Across All Domains}} \\
\textsc{age}          & \multicolumn{3}{c}{Age / Date of Birth} \\
\textsc{occupation}   & \multicolumn{3}{c}{Occupation / Job Title} \\
\textsc{location}     & \multicolumn{3}{c}{City / Region / Zip} \\
\textsc{affiliation}  & \multicolumn{3}{c}{Employer / Organization} \\
\textsc{family}       & Family members & Dependents & Heirs \\
\midrule
\multicolumn{4}{c}{\textbf{Domain-Specific Classes}} \\
\addlinespace
Specific 1 & Condition      & Asset Class    & Matter Type \\
Specific 2 & Treatment      & Income Band    & Proc. Role \\
Specific 3 & Visit Pattern  & Transactions    & Case Disp. \\
\bottomrule
\end{tabular}
\end{table}

\begin{itemize}
    \item \textit{QI Distribution:} Recognizing that single identifiers rarely facilitate re-identification, we sample $k \in \{2,3,4\}$ QIs per unsafe record with a prior distribution of $(0.45, 0.35, 0.20)$, biasing toward the cluster sizes most dominant in re-identification literature.

    \item \textit{Direct-Identifier Prohibitions:} Our validator rejects records containing direct identifiers. Universal prohibitions include SSNs, contact information, and formatted dates of birth. Domain-specific constraints include Medical Record Numbers (MRNs) for medicine; credit card and account numbers for finance; and docket or bar numbers for law.
\end{itemize}

\subsection{Generation Axes}
\label{sec:data:axes}

For each unsafe record, we sample independently from four axes so that
coverage scales combinatorially rather than by template enumeration:

\begin{itemize}
    \item \textit{QI type (8 levels)}  Specifies which quasi-identifier dimensions are populated. (Table~\ref{tab:qimap}).
    \item \textit{Framing (6 levels)}  Defines the discourse register, such as \textsc{case\_voice}, \textsc{implicit\_case}, or \textsc{referral}.
    \item \textit{Placement (4 levels)} : Determines the position of the QI cluster (e.g., \textsc{frame\_opening} or \textsc{distributed}).
    \item \textit{Adversarial mode (4 levels)}: Dictates the leakage style, ranging from literal mentions (none) to paraphrased \textsc{indirect\_qi} or \textsc{style\_transfer}.
\end{itemize}

This combinatoric approach ensures that no single stylistic shortcut can dominate the unsafe class, forcing the detector to learn privacy semantics rather than simple patterns. 

\subsection{Cross-Generator Diversification and Validation}
\label{sec:data:diversity}

We utilize three open-weight generators (Qwen2.5-7B, Mistral-7B, and Phi-3-medium) to populate the training pool, while holding out a fourth lineage (Yi-1.5-9B) for cross-generator evaluation. (Section~\ref{sec:eval:crossgen}). Each generator runs on a vLLM-served OpenAI-compatible endpoint at temperature $0.9$ with max\_tokens $600$ and max\_model\_len $4096$.

For each domain, each training generator produces three batches: a $15{,}000$-record baseline; $1{,}500$ records of \textsc{indirect\_qi}; and $3{,}000$ records of mixed \textsc{distractor\_padded} + \textsc{style\_transfer}. Yi-1.5-9B contributes a single $2{,}000$-record baseline batch held out from training. Aggregate per-domain totals are shown in Table~\ref{tab:gen}.


\begin{table}[t]
\centering
\caption{Generation-campaign totals across the three domains. Each
domain uses three training-pool generators (Qwen2.5-7B, Mistral-7B-v0.3,
Phi-3-medium-4k) plus one held-out generator (Yi-1.5-9B). Per-generator
breakdowns are deferred to Appendix~\ref{app:gen}.}
\label{tab:gen}
\small
\begin{tabular}{lrrr}
\toprule
\textbf{Set} & \textbf{Medical} & \textbf{Finance} & \textbf{Law} \\
\midrule
Safe seed corpus       & 46{,}412 & 42{,}405 & 34{,}065 \\
Unsafe (training pool) & 47{,}916 & 58{,}500 & 58{,}500 \\
Unsafe (Yi held out)   &  2{,}000 &  2{,}000 &  2{,}000 \\
Borderline aug (Mistral) & 4{,}000 & 4{,}000 & 4{,}000 \\
Borderline eval (Phi-3) & 1{,}000 & 1{,}000 & 1{,}000 \\
\midrule
\textbf{Total generated} & \textbf{54{,}916} & \textbf{65{,}500} & \textbf{65{,}500} \\
\bottomrule
\end{tabular}
\end{table}

Each generated record undergoes validation through a domain-specific validator that rejects records containing direct identifiers, named individuals, formatted DOBs, or, in adversarial-mode runs, vocabulary that the prompt explicitly forbade. Rejected records are retried (with a per-mode retry budget) against a fresh seed/spec/model triple; records that exhaust retries are dropped.

\subsection{Borderline-Safe Construction}
\label{sec:data:borderline}

A dual density detector mechanism evaluated only on within-distribution holdouts cannot distinguish between learning the QI-leakage signal and learning the referential-voice surface feature. The diagnostic finding of this paper (Section~\ref{sec:eval:borderline}) requires a \emph{borderline-safe} corpus: genuinely-safe content stylistically matched to the unsafe class but containing zero QI clusters. This corpus is the negative class in the central stress test and the training augmentation that defines the v4 detector configuration. The corpus spans five subtypes, sampled uniformly:

\begin{itemize}
    \item \textit{Population Demographics:} Cohort-level statements (e.g., "adults aged 50–70") without individual framing.
    \item \textit{Public Guideline Quotes:} Lengthy passages from authoritative sources or attributed quote from a publicly available clinical guideline (CDC, AHA, USPSTF, WHO, NICE). Mimics verbatim-source leakage but with public, non-patient source material.
    \item \textit{Second-Person Education:} Purely instructional second person voice.
    \item \textit{Anonymized Case Reports:} Case-study narratives stripped of all demographic or QI detail.
    \item \textit{Empathetic Deflection:} Standard consultative redirections.
\end{itemize}

To ensure evaluation integrity, we maintain two disjoint borderline sets generated by different models (Mistral for training, Phi-3 for testing). This separation ensures that improvements in False Positive Rates (FPR) reflect true semantic learning rather than the memorization of a specific generator's style.

\section{Detection Framework}
\label{sec:detection}

\begin{figure}[]
    \centering
    \includegraphics[width=\columnwidth]{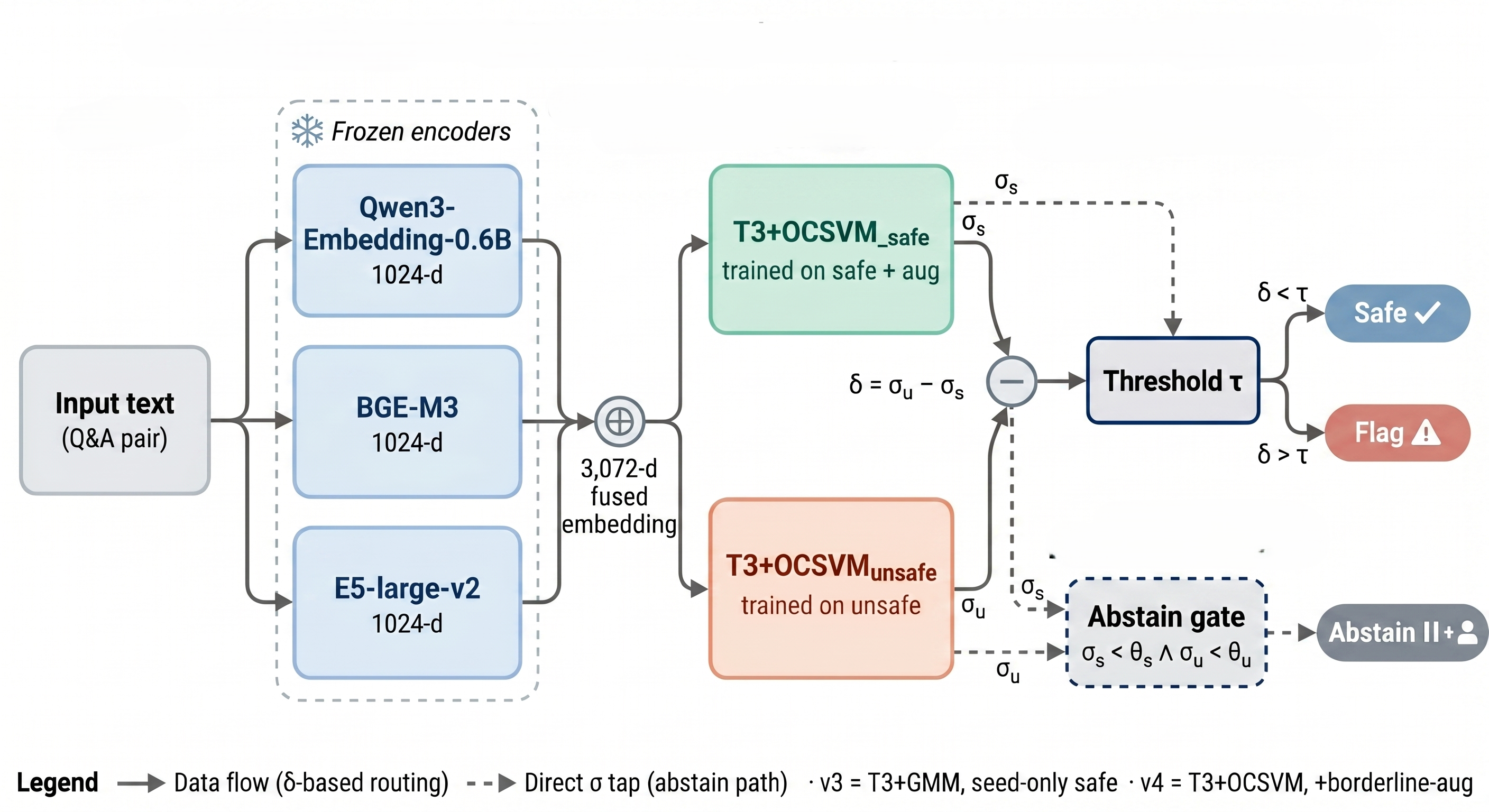}
    \caption{Layer-2 detector architecture. Three frozen encoders
    produce a fused $3{,}072$-d representation; two one-class
    detectors fit in parallel ($\sigma_s$ on safe + borderline-aug,
    $\sigma_u$ on unsafe). The discriminator
    $\delta = \sigma_u - \sigma_s$ is thresholded by $\tau$; the
    abstain gate routes both OOD inputs to human review
    (Section~\ref{sec:disc:operating}).}
    \label{fig:detector}
\end{figure}

The Layer~2 detector is designed as a dual one-class density estimator operating over a fused text-embedding representation of the generated answer. This architecture (Figure~\ref{fig:detector}) deliberately avoids learned classification heads, ensuring that training requires only the safe and unsafe exemplar corpora defined in Section~\ref{sec:data}. Detection consists of a single forward pass through frozen encoders followed by parallel density evaluations. We describe the embedding stack and density estimators below, then formalize the discriminator and the principled abstain gate.

\subsection{Embedding Stack}
\label{sec:detection:embed}

Each Q\&A pair is encoded by three frozen text encoders, with the resulting representations concatenated into a single feature vector:

\begin{itemize}
    \item \textbf{Qwen3-Embedding-0.6B} (Alibaba): Provides a dense semantic representation ($1024$-dim).
    \item \textbf{BGE-M3} (BAAI): multilingual dense retrieval
          embedding ($1024$-dim).
    \item \textbf{E5-large-v2} (Microsoft): general-purpose retrieval
          embedding ($1024$-dim).
\end{itemize}

The combined $3{,}072$-dimensional representation serves as input for both one-class detectors. We employ multiple encoders because the "case-style" signal covers surface phrasing, semantic content, and register. A single encoder usually fails to capture all these aspects effectively. By concatenating features instead of ensembling at the end, the density estimators can directly learn cross-encoder relationships without needing an extra model to merge scores.

\subsection{One-Class Density Estimators}
\label{sec:detection:density}

We train two one-class detectors in parallel:

\begin{itemize}
    \item $\sigma_s$: Trained on the safe corpus. In configuration \textit{v3}, this uses the seed corpus alone; in \textit{v4}, it is augmented with the $4{,}000$-record borderline-safe set.
    \item $\sigma_u$: a one-class detector trained on the unsafe corpus. This training set remains consistent across both \textit{v3} and \textit{v4} configurations.
\end{itemize}

We evaluate two density-estimator choices: a Gaussian Mixture Model
(T3+GMM, used in v3) and a one-class SVM (T3+OCSVM, used in v4). Our ablation study (Section~\ref{sec:eval:ablation_analysis}) indicates that T3+OCSVM is significantly more robust to borderline-safe inputs. T3+GMM components are fit with $K{=}8$, full covariance PCA-reduced ($512$-dim) features. T3+OCSVM utilizes an RBF kernel with $\gamma$ set via the median heuristic and $\nu$ selected via a 5-fold validation sweep over $\{0.005, 0.01, 0.02, 0.05\}$ (e.g., $\nu = 0.01$ for the law domain).

\subsection{Discriminator and Operational Thresholds}
\label{sec:detection:discrim}

At inference time the detector emits the discriminator score
\[
  \delta(x) \;=\; \sigma_u(x) \;-\; \sigma_s(x),
\]
where a positive $\delta(x)$ indicates the input lies closer to the unsafe distribution than the safe one. A flag is emitted iff $\delta(x) > \tau$ for an operator-chosen threshold $\tau$. The \textit{T3+OCSVM-v4} detector supports two primary operating points:

\begin{itemize}
    \item \textbf{Conservative ($\tau = 0$).} Flag iff
          $\sigma_u(x) > \sigma_s(x)$. Low FPR, lower TPR; suitable
          where false-positive cost dominates (clinician-facing tools,
          attorney-facing productivity tools).
    \item \textbf{Balanced ($\tau$ at FPR@90\%TPR).} Threshold chosen
          so the within-distribution TPR is $0.90$.
    \item \textbf{Strict ($\tau$ at FPR@95\%TPR).} Threshold chosen
          so the within-distribution TPR is $0.95$. Higher FPR, higher
          TPR; suitable for upstream sanitization filters where
          missed leaks are the dominant cost.
\end{itemize}

\subsection{Two-Distribution OOD and the Abstain Gate}
\label{sec:detection:abstain}

The discriminator $\delta$ is well-calibrated only when at least one
of $\sigma_s(x)$ or $\sigma_u(x)$ claims the input as in-distribution.
The discriminator $\delta$ is numerically unstable when an input is a "tail value" for both distributions (i.e., it looks like neither safe nor unsafe data). In such cases, the difference between likelihoods is dominated by estimator noise rather than privacy signals.

To address this, we implement a principled abstain region: a gate defined by $\sigma_s(x) < \theta_s \wedge \sigma_u(x) < \theta_u$. Inputs falling into this region are routed to human review or a secondary detection layer. While the T3+OCSVM-v4 configuration naturally compresses this regime compared to the T3+GMM baseline (Section~\ref{sec:disc:operating}), the inclusion of an abstain gate remains a critical methodological requirement for robust deployment.

\section{Evaluation}
\label{sec:eval}

We evaluate our dual detector mechanism across three domains and three distinct evaluation pairings. Our headline finding is that the case-style confound exposed by the borderline-safe stress test (Section~\ref{sec:eval:borderline}) generalizes across all three domains, as does the effectiveness of the T3+OCSVM-v4. While the remediation is portable, we identify a unique residual structure in the finance domain, which we discuss in Section~\ref{sec:disc:generalization}.

\subsection{Evaluation Framework and Metrics}
\label{sec:setup:configs}

To ensure clarity, we strictly distinguish between the detector \textit{Configuration} and the \textit{Test Pairing} as follows:

\begin{itemize}
    \item \textit{Configurations (Training Property):}

            \begin{itemize}
                \item \textit{v3}: A safe-side detector trained exclusively on the seed corpus.
                \item \textit{v4}: A safe-side detector trained on the seed corpus augmented with the $4{,}000$-record borderline-safe dataset.
                \item \textit{T3+GMM-v3} and \textit{T3+OCSVM-v4} represent the baseline and recommended deployment configurations, respectively.
            \end{itemize}
    
    \item \textit{Test pairing (Evaluation Property):} 
          \begin{itemize}
              \item \emph{Within-Distribution:} Unsafe vs. safe holdouts from the training distributions.
              \item \emph{Cross-Generator:} Evaluates generalization by testing against unsafe records produced by \textit{Yi-1.5-9B-Chat} (a 9B-class generator from a distinct lineage, held out from unsafe-side training in all three domains). The goal is to test whether the detector learned the privacy-leakage signal or merely the stylometric fingerprint of the three training-pool generators \textit{(Qwen2.5-7B, Mistral-7B-v0.3, Phi-3-medium-4k)}.
              \item \emph{Borderline-Safe Stress Test:} Pairs unsafe holdouts against a $1{,}000$-record corpus of genuinely safe content that matches the ``client-voice'' style of the unsafe class.
          \end{itemize}
\end{itemize}

The v3/v4 distinction is therefore about \emph{training data}; the within-dist / cross-gen/borderline distinction is about \emph{test data}. A given configuration (say T3+OCSVM-v4) is evaluated against all three test pairings, producing one row in each of three evaluation tables (see Tables~\ref{tab:borderline},~\ref{tab:borderline:strata} and \ref{tab:crossgen}).

All reported binary metrics (FPR, TPR, AUROC, FPR$_{95}$, FPR$_{90}$) are computed on the \emph{kept population}: records that the abstain gate (Section~\ref{sec:detection:abstain}) would have routed to human review are excluded from both the numerator and the denominator. Abstain rate (Table~\ref{tab:abstain}) and binary metrics are therefore reported as independent measurements of two different deployment-time decisions: \emph{which records the system commits to} (abstain rate) and \emph{how well it decides on the records it commits to} (FPR/TPR). Threshold-calibrated quantities (FPR$_{95}$, FPR$_{90}$) are read off the ROC of the kept population. The abstain thresholds $(\theta_s, \theta_u)$ are calibrated once per detector configuration on the within-distribution holdout, then applied identically to all three test pairings; this matches the operational deployment in which abstain thresholds are fixed at training time.

\subsection{Within-Distribution Detection}
\label{sec:eval:withindist}

Table~\ref{tab:withindist} summarizes detection metrics for the within-distribution holdout pairings across the three domains. While both configurations achieve high AUROC, T3+OCSVM-v4 significantly optimizes the $\text{FPR}_{95}$ operating point, most notably in the law domain where error rates fall from $42.2\%$ to $18.5\%$. We report $\text{FPR}_{90}$ alongside $\text{FPR}_{95}$ to provide a more lenient deployment perspective, as the $95\%$ TPR target is a stringent operating point that often magnifies ROC-tail noise. Across all domains, a modest $5$\,pp TPR concession yields substantial FPR headroom, with the $\text{FPR}_{95} \to \text{FPR}_{90}$ ratio holding at approximately $2\text{--}3\times$ (e.g., law T3+OCSVM-v4: $18.5\% \to 7.2\%$).

The law domain's within-distribution AUROC remains lower than medical and finance, a disparity we attribute to the law safe corpus spanning five sources across four jurisdictions. However, T3+OCSVM closes this gap more effectively than T3+GMM, corroborating the estimator-choice findings in Section~\ref{sec:eval:ablation_analysis}. Crucially, while these within-distribution metrics might superficially justify the production deployment of T3+GMM-v3 ($\text{AUROC} \geq 0.93$), the borderline-safe stress test (Section~\ref{sec:eval:borderline}) reveals that such performance can be misleading.

\subsection{Cross-Generator Generalization}
\label{sec:eval:crossgen}

A central concern in synthetic training is that detectors may overfit to generator-specific stylometry rather than capturing the underlying privacy-leakage signal. Table~\ref{tab:crossgen} addresses this by pairing unsafe records from the held-out Yi-1.5-9B-Chat model against the safe holdout under T3+OCSVM-v4. Across all domains, the held-out-generator AUROC remains $\geq 0.95$ (kept population). In the medical and finance sectors, cross-generator performance effectively matches or exceeds within-distribution benchmarks; in the law domain, the slight decrease ($0.9580$ vs.\ $0.9711$) is likely due to the law safe corpus's inherent breadth and the Yi model's stylistic distance from the training pool.

Analysis of the FPR metrics in Table~\ref{tab:crossgen} further localizes this law-domain gap. While medical and finance cross-generator FPRs match or beat their within-distribution counterparts, the law domain shows a specific divergence at the $\text{FPR}_{90}$ level: the cross-gen value ($13.4\%$) is $1.9\times$ higher than the within-distribution value ($7.2\%$), despite $\text{FPR}_{95}$ remaining nearly equal. This suggests the performance gap resides in the bulk of the score distribution rather than the extreme ROC tail. As confirmed by per-generator stratification (Appendix~\ref{app:strata}), Yi-on-law TPR at $\tau{=}0$ is $0.51$ compared to the training-pool mean of $0.72$---a significantly wider disparity than the $\leq 3 pp$ gap seen in other domains. 

Ultimately, these results demonstrate that T3+OCSVM-v4 successfully avoids single-generator stylometric overfitting within the 7--14B instruction-tuned model class. However, the efficacy of this approach against closed-source frontier models or gradient-guided adversarial generation remains an area for future investigation (Section~\ref{sec:limitations}).

\subsection{The Borderline-Safe Stress Test}
\label{sec:eval:borderline}

The borderline-safe stress test serves as the central diagnostic tool, distinguishing whether the detector captures the intended quasi-identifier (QI) signal or merely identifies "referential client voice"—a stylistic confound prevalent in synthetic training distributions. By pairing the unsafe holdout with $1{,}000$-record Phi-3 records that are stylistically identical to the unsafe class but contain zero QI clusters, we expose a significant "case-style confound" across all domains. While T3+GMM-v3 achieves high within-distribution AUROC ($0.95$--$1.00$), its borderline performance collapses to $[0.72, 0.79]$. This non-overlapping range across three independent corpora and taxonomies suggests that the diagnostic captures an inherent property of the dual one-class density paradigm rather than a data-specific artifact.

The T3+OCSVM-v4 ``combined-fix'' recipe successfully remediates this gap, recovering borderline AUROC to $\geq 0.93$ in all domains and reducing $\text{FPR}_{95}$ by $44$--$55$\,pp depending on domain. Comparable reductions at the more lenient $\text{FPR}_{90}$ operating point ($43$--$56$\,pp) confirm that the improvement is not a local artifact of a single TPR target but a shift of the entire ROC tail. Notably, per-subtype stratification (Table~\ref{tab:borderline:strata}) shows that while medical and law converge to a $\sim 4\%$ mean residual, finance retains a $\sim 35\%$ residual. This is localized to quantitative subtypes such as \textsc{second\_person\_education} and \textsc{population\_demographics} where cohort-scoped data (e.g., income bands) is stylistically indistinguishable from single-client QI signals in the embedding space.
Another reason for the finance FPR being high at $\tau=0$ compared to other domains is cross-domain operating-point non-comparability. Reading FPR at $\tau=0$ across domains is partly misleading because $\tau=0$ lands at different effective TPRs in each domain's score distribution: 80.2\% (medical), 97.8\% (finance), 72.0\% (law). Finance's $\tau=0$ sits $\sim18pp$ deeper into the recall-greedy regime than medical's, and its higher $\tau=0$ FPR therefore partly reflects this calibration difference rather than a domain-specific detection deficit. We retain $\tau=0$ reporting because the within-domain v3-vs-v4 contrast (the binding comparison for the case-style confound) shares a domain-specific score distribution and is well-posed; cross-domain comparisons in this paper use FPR95 and FPR90 wherever a fair comparison is the binding requirement.

\subsection{Operational Feasibility: The Abstain Rate}
\label{sec:eval:abstain}

The second critical benefit of T3+OCSVM-v4 is the stabilization of the abstain gate, which routes records to human review when neither density ($\sigma_s, \sigma_u$) can confidently claim the input. Under T3+GMM-v3, the medical and finance domains exhibit "borderline abstain rates" of $31.6\%$ and $27.5\%$, respectively. Such figures represent an operationally infeasible human-review load at scale. T3+OCSVM-v4 reduces these rates by $3\times$--$8\times$: medical drops to $10.7\%$, while finance and law fall to $\leq 3.7\%$. 

T3+GMM-v3 saturates the review queue because its log-likelihoods bottom out on stylistically novel content; once both densities saturate, the discriminator becomes dominated by tail noise, triggering the abstain gate correctly but far too frequently. Law is the sole exception (T3+GMM-v3 abstain at $4.6\%$), likely because its heterogeneous safe corpus (five sources across four jurisdictions) creates a wider safe distribution with fewer deep-tail records. Overall, T3+OCSVM-v4 proves to be the only configuration that simultaneously fixes binary error rates and maintainable review-queue volumes, making it the recommended choice for production deployment.

\subsection{Ablation: Why T3+OCSVM Outperforms T3+GMM}
\label{sec:eval:ablation_analysis}

Isolating the estimator choice reveals that T3+OCSVM consistently outperforms T3+GMM under stylistic overlap. In the medical ablation grid, T3+OCSVM matches T3+GMM on within-distribution metrics but provides a $0.18$ AUROC boost on the borderline-safe stress test under matched v3 data. This disparity is structural: a T3+GMM partitions feature space into Gaussian basins where log-likelihoods are dominated by the distance to the nearest cluster mean. Genuinely safe content that falls between these clusters receives extreme negative log-likelihoods, saturating the safe density and causing the discriminator to misclassify the input as unsafe.

In contrast, T3+OCSVM defines a singular boundary around the support of the training distribution and assigns graduated signed-distance scores. Because points outside the boundary receive linear distance scores rather than exponential tail values, the discriminator remains numerically well-conditioned even for inputs that are technically out-of-distribution (OOD). This result validates T3+OCSVM as the superior estimator for high-stakes privacy detection where the boundary between "safe style" and "unsafe content" is inherently porous.

\begin{table}[h]
\centering
\caption{Abstain rate per (domain, configuration, test set) under
the $5$th-percentile rule of
Section~\ref{sec:detection:abstain}. Borderline column is the
deployability signal; T3+OCSVM-v4 reduces it $3$--$8\times$ relative
to T3+GMM-v3.}
\label{tab:abstain}
\footnotesize
\setlength{\tabcolsep}{5pt}
\begin{tabular}{llccc}
\toprule
\textbf{Domain} & \textbf{Cfg}
  & \textbf{Safe holdout} & \textbf{Unsafe corpus} & \textbf{Borderline} \\
\midrule
Medical  & T3+GMM-v3   & $4.45\%$ & $4.61\%$ & $31.60\%$ \\
Medical  & \textbf{T3+OCSVM-v4} & $4.44\%$ & $4.35\%$ & $\mathbf{10.70\%}$ \\
\midrule
Finance  & T3+GMM-v3   & $3.70\%$ & $4.35\%$ & $27.50\%$ \\
Finance  & \textbf{T3+OCSVM-v4} & $4.30\%$ & $3.54\%$ & $\mathbf{3.50\%}$ \\
\midrule
Law      & T3+GMM-v3   & $0.87\%$ & $4.40\%$ & $4.60\%$  \\
Law      & \textbf{T3+OCSVM-v4} & $2.42\%$ & $3.83\%$ & $\mathbf{3.70\%}$ \\
\bottomrule
\end{tabular}
\end{table}

\begin{table*}[t]
\centering
\caption{Within-distribution detection metrics: unsafe vs.\ safe
holdout, per domain. v3/v4 refer to safe-side training data
(Section~\ref{sec:setup:configs}). FPR$_{95}$ / FPR$_{90}$ are FPR
at $95\%$ / $90\%$ TPR.}
\label{tab:withindist}
\footnotesize
\setlength{\tabcolsep}{3pt}
\begin{tabular}{lccccc|ccccc|ccccc}
\toprule
& \multicolumn{5}{c}{\textbf{Medical}}
& \multicolumn{5}{c}{\textbf{Finance}}
& \multicolumn{5}{c}{\textbf{Law}} \\
\textbf{Config}
  & AUROC & FPR$_{95}$ & FPR$_{90}$ & TPR$_{\tau0}$ & FPR$_{\tau0}$
  & AUROC & FPR$_{95}$ & FPR$_{90}$ & TPR$_{\tau0}$ & FPR$_{\tau0}$
  & AUROC & FPR$_{95}$ & FPR$_{90}$ & TPR$_{\tau0}$ & FPR$_{\tau0}$ \\
\midrule
T3+GMM-v3       & 0.995 &  2.7\% & 1.1\% & 98.0\% &  6.3\%
             & 0.984 & 10.6\% & 4.3\% & 96.1\% & 12.9\%
             & 0.946 & 42.2\% & 18.0\% & 93.3\% & 32.6\% \\
\textbf{T3+OCSVM-v4}
             & \textbf{0.995} & \textbf{2.5\%} & \textbf{0.7\%} & \textbf{80.2\%} & \textbf{0.1\%}
             & \textbf{0.992} & \textbf{3.6\%} & \textbf{1.6\%} & \textbf{97.8\%} & \textbf{9.7\%}
             & \textbf{0.971} & \textbf{18.5\%} & \textbf{7.2\%} & \textbf{72.0\%} & \textbf{0.7\%} \\
\bottomrule
\end{tabular}
\end{table*}

\begin{table*}[t]
\centering
\caption{Borderline-safe stress test: unsafe holdout vs.\ the
$1{,}000$-record Phi-3 borderline-safe corpus
(Section~\ref{sec:data:borderline}). TPR$_{\tau0}$ is omitted; the
unsafe set is identical to Table~\ref{tab:withindist}'s.}
\label{tab:borderline}
\footnotesize
\setlength{\tabcolsep}{4pt}
\begin{tabular}{lcccc|cccc|cccc}
\toprule
& \multicolumn{4}{c}{\textbf{Medical}}
& \multicolumn{4}{c}{\textbf{Finance}}
& \multicolumn{4}{c}{\textbf{Law}} \\
\textbf{Config}
  & AUROC & FPR$_{95}$ & FPR$_{90}$ & FPR$_{\tau0}$
  & AUROC & FPR$_{95}$ & FPR$_{90}$ & FPR$_{\tau0}$
  & AUROC & FPR$_{95}$ & FPR$_{90}$ & FPR$_{\tau0}$ \\
\midrule
T3+GMM-v3       & 0.725 & 77.6\% & 64.9\% & 86.7\%
             & 0.722 & 74.2\% & 64.3\% & 77.9\%
             & 0.792 & 85.3\% & 65.7\% & 78.3\% \\
\textbf{T3+OCSVM-v4}
             & \textbf{0.965} & \textbf{22.8\%} & \textbf{9.0\%} & \textbf{2.4\%}
             & \textbf{0.963} & \textbf{21.8\%} & \textbf{11.2\%} & \textbf{36.0\%}
             & \textbf{0.930} & \textbf{41.0\%} & \textbf{22.4\%} & \textbf{3.1\%} \\
\midrule
$\Delta$AUROC (v3$\to$v4) & \multicolumn{4}{c}{$+0.241$} & \multicolumn{4}{c}{$+0.241$} & \multicolumn{4}{c}{$+0.139$} \\
$\Delta$FPR$_{95}$        & \multicolumn{4}{c}{$-54.8$\,pp} & \multicolumn{4}{c}{$-52.5$\,pp} & \multicolumn{4}{c}{$-44.3$\,pp} \\
$\Delta$FPR$_{90}$        & \multicolumn{4}{c}{$-55.9$\,pp} & \multicolumn{4}{c}{$-53.1$\,pp} & \multicolumn{4}{c}{$-43.3$\,pp} \\
\bottomrule
\end{tabular}
\end{table*}

\begin{table*}[t]
\centering
\caption{Per-subtype borderline FPR at $\tau{=}0$, broken out
across the five borderline-safe subtypes
(Section~\ref{sec:data:borderline}). Per-subtype rows and
\textbf{Mean (subtype)} are full-population (kept-only per-subtype
counts are too small for stable estimates). \textbf{Global
FPR$_{\tau0}$} uses the kept-only convention of
Table~\ref{tab:borderline} for direct comparison; per-domain
analysis is in Section~\ref{sec:disc:generalization}.}
\label{tab:borderline:strata}
\footnotesize
\setlength{\tabcolsep}{5pt}
\begin{tabular}{lcc|cc|cc}
\toprule
& \multicolumn{2}{c}{\textbf{Medical}}
& \multicolumn{2}{c}{\textbf{Finance}}
& \multicolumn{2}{c}{\textbf{Law}} \\
\textbf{Subtype} & v3 & v4 & v3 & v4 & v3 & v4 \\
\midrule
\textsc{anonymized\_case\_report}  & 97.5\% & 13.1\% & 93.5\% & 26.9\% & 80.8\% &  3.4\% \\
\textsc{empathetic\_deflection}    & 66.7\% &  0.0\% & 70.8\% & 18.1\% & 68.0\% &  0.0\% \\
\textsc{population\_demographics}  & 89.5\% &  5.0\% & 91.9\% & 47.7\% & 92.6\% & 18.0\% \\
\textsc{public\_guideline\_quote}  & 87.0\% &  2.2\% & 58.3\% & 35.0\% & 72.8\% &  0.5\% \\
\textsc{second\_person\_education} & 88.4\% &  2.0\% & 72.3\% & 49.3\% & 74.2\% &  0.5\% \\
\midrule
\textbf{Mean (subtype)}          & 85.8\% & \textbf{4.5\%} & 77.4\% & \textbf{35.4\%} & 77.7\% & \textbf{4.5\%} \\
\textbf{Global FPR$_{\tau0}$ (kept)} & 86.7\% & \textbf{2.4\%} & 77.9\% & \textbf{36.0\%} & 78.3\% & \textbf{3.1\%} \\
\bottomrule
\end{tabular}
\end{table*}

\begin{table}[t]
\centering
\caption{Cross-generator generalization: unsafe records from
held-out Yi-1.5-9B-Chat (distinct lineage) vs.\ the same safe
holdout as Table~\ref{tab:withindist}, T3+OCSVM-v4 only.
Within-distribution operating points repeated for direct
comparison.}
\label{tab:crossgen}
\small
\begin{tabular}{lccc}
\toprule
\textbf{Metric (T3+OCSVM-v4)} & \textbf{Medical} & \textbf{Finance} & \textbf{Law} \\
\midrule
Within-distribution AUROC      & 0.9953 & 0.9920 & 0.9711 \\
Cross-generator AUROC          & 0.9974 & 0.9965 & 0.9580 \\
$\Delta$AUROC vs.\ within-dist & $+0.002$ & $+0.005$ & $-0.013$ \\
\midrule
Cross-gen FPR$_{95}$           & 1.6\%  & 1.8\%  & 19.5\% \\
\quad (within-dist FPR$_{95}$) & 2.5\%  & 3.6\%  & 18.5\% \\
Cross-gen FPR$_{90}$           & 0.7\%  & 0.9\%  & 13.4\% \\
\quad (within-dist FPR$_{90}$) & 0.7\%  & 1.6\%  & 7.2\%  \\
\bottomrule
\end{tabular}
\end{table}

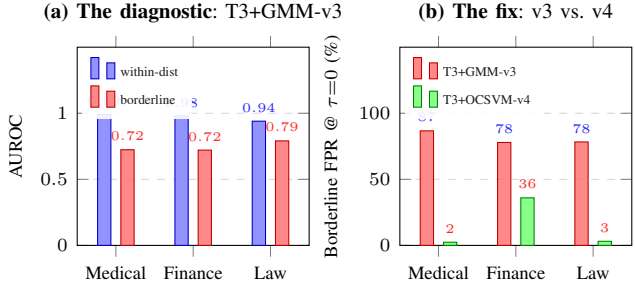
\begin{figure}[]
\centering
\pgfplotsset{
  resbar/.style={
    ybar=3.75pt,
    bar width=5pt, 
    width=0.525\columnwidth,   
    height=4.2cm,  
    enlarge x limits=0.25,
    ymajorgrids,
    grid style={dashed, gray!30},
    legend style={at={(0.02,0.98)}, anchor=north west, font=\tiny,
                  draw=none, fill=white, fill opacity=0.9},
    legend cell align=left,
    symbolic x coords={Medical, Finance, Law},
    xtick=data,
    tick label style={font=\scriptsize},
    label style={font=\scriptsize},
    every axis plot/.append style={fill opacity=0.85, draw opacity=1},
  },
}

\begin{tikzpicture}
\begin{groupplot}[
  group style={
    group size=2 by 1, 
    horizontal sep=12mm, 
    x descriptions at=edge bottom
  },
  resbar,
]

\nextgroupplot[
  title={\footnotesize\textbf{(a) The diagnostic}: T3+GMM-v3},
  ylabel={AUROC},
  ymin=0.0, ymax=1.5,
  ytick={0.0, 0.5, 1.0},
  nodes near coords, nodes near coords style={font=\tiny,
    /pgf/number format/fixed, /pgf/number format/precision=2},
]
\addplot+[fill=blue!50,  draw=blue!75!black] coordinates
  {(Medical, 0.99) (Finance, 0.98) (Law, 0.94)};
\addplot+[fill=red!55,   draw=red!75!black]  coordinates
  {(Medical, 0.724) (Finance, 0.721) (Law, 0.791)};
\legend{within-dist, borderline}

\nextgroupplot[
  title={\footnotesize\textbf{(b) The fix}: v3 vs.\ v4},
  ylabel={Borderline FPR @ $\tau{=}0$ (\%)},
  ymin=0, ymax=150,
  ytick={0, 50, 100},
  nodes near coords, nodes near coords style={font=\tiny,
    /pgf/number format/fixed, /pgf/number format/precision=0},
]
\addplot+[fill=red!55,    draw=red!75!black]    coordinates
  {(Medical, 86.7) (Finance, 77.9) (Law, 78.3)};
\addplot+[fill=green!55,  draw=green!50!black]  coordinates
  {(Medical, 2.4) (Finance, 36.0) (Law, 3.1)};
\legend{T3+GMM-v3, T3+OCSVM-v4}
\end{groupplot}
\end{tikzpicture}
\caption{Case-style confound and its remediation. (a) AUROC performance per domain. (b) Comparison of False Positive Rates (FPR) at $\tau{=}0$. Finance's elevated FPR in (b) reflects calibration differences discussed in Section~\ref{sec:eval:borderline}.}
\label{fig:results}
\end{figure}

\section{Comparative Analysis: Architecture Alternatives}
\label{sec:comparison}

We benchmark the proposed one-class detector against two alternatives when forced to commit, those same records cross share the same evaluation substrate: (i)~a supervised binary classifier (MLP-v4) sharing the same encoder backbone, trained with cross-entropy on safe$=0$/unsafe$=1$ labels; and (ii)~an LLM-as-judge baseline Qwen2.5-14B-Instruct served via vLLM, prompted with a QI definition and 3 SAFE\,/\,3 UNSAFE few-shot examples per domain). Both alternatives use the same train/test splits as OCSVM-v4 and write per-record scores in the schema of Table~\ref{tab:withindist}. Table~\ref{tab:baselines} reports the comparison on the borderline-safe pairing. We discuss our main findings in the following.

\begin{table*}[t]
\centering
\caption{Three architectures on the borderline-safe stress test. Tuples report medical / finance / law. Bold marks the per-row optimum; \textcolor{red}{red} marks orders-of-magnitude losses that disqualify a baseline from operational deployment.}
\label{tab:baselines}
\footnotesize
\setlength{\tabcolsep}{5pt}
\begin{tabular}{lccc}
\toprule
\textbf{Metric}
  & \textbf{OCSVM-v4 (ours)}
  & \textbf{MLP-v4 (binary)}
  & \textbf{Qwen-14B (LLM-judge)} \\
\midrule
\multicolumn{4}{l}{\emph{Operational footprint --- how much of the borderline traffic the system handles}} \\
\addlinespace[2pt]
\textit{(borderline pairing)}
  & \scriptsize\textit{med~/~fin~/~law}
  & \scriptsize\textit{med~/~fin~/~law}
  & \scriptsize\textit{med~/~fin~/~law} \\
Abstain rate (records routed to review)
  & \textbf{10.7\% / 3.5\% / 3.7\%}
  & \textcolor{red}{85.0\% / 72.8\% / 69.7\%}
  & n/a\textsuperscript{$\dagger$} \\
\midrule
\multicolumn{4}{l}{\emph{With abstain on (kept-population at $\tau{=}0$): metrics on records the system commits to}} \\
\addlinespace[2pt]
\textit{(borderline pairing)}
  & \scriptsize\textit{med~/~fin~/~law}
  & \scriptsize\textit{med~/~fin~/~law}
  & \scriptsize\textit{med~/~fin~/~law} \\
AUROC
  & 0.97 / 0.96 / 0.93
  & \textbf{1.00 / 1.00 / 0.99}
  & 0.92 / 0.94 / 0.93 \\
FPR (false-flag on safe)
  & \textbf{2.4\%} / 36.0\% / \textbf{3.1\%}
  & 2.7\% / \textbf{5.5\%} / 12.5\%
  & 30.2\% / 8.6\% / 5.4\% \\
$n_\text{kept}$ (out of 1{,}000 borderline)
  & 893 / 965 / 963
  & \textcolor{red}{150 / 272 / 303}
  & 1000 / 1000 / 1000 \\
\midrule
\multicolumn{4}{l}{\emph{With abstain off (full-population at $\tau{=}0$, MLP forced to commit on every input)}} \\
\addlinespace[2pt]
\textit{(borderline pairing)}
  & \scriptsize\textit{med~/~fin~/~law}
  & \scriptsize\textit{med~/~fin~/~law}
  & \scriptsize\textit{med~/~fin~/~law} \\
FPR (false-flag on safe)
  & \textbf{4.0\%} / 35.1\% / \textbf{4.3\%}
  & \textcolor{red}{28.3\% / 47.1\% / 26.2\%}
  & 30.2\% / 8.6\% / 5.4\% \\
\midrule
\multicolumn{4}{l}{\emph{Operations}} \\
Per-query latency / cost
  & \textbf{${\sim}5$\,ms / ${\sim}\$10^{-5}$}
  & \textbf{${\sim}5$\,ms / ${\sim}\$10^{-5}$}
  & \textcolor{red}{${\sim}1$\,s / ${\sim}\$10^{-2}$} \\
Calibratable to operating point
  & \textbf{yes}
  & \textbf{yes}
  & no (discrete output) \\
\bottomrule
\end{tabular}
\\[0.4em]
\textsuperscript{$\dagger$}LLM-judge has no calibrated abstain mechanism.
\end{table*}

\paragraph{The MLP binary classifier has no operationally viable operating point.} Table~\ref{tab:baselines} makes the trade-off explicit. With the abstain mechanism enabled, MLP-v4 achieves competitive per-record FPR ($2.7$--$12.5\%$ on the records it
commits to), but only because it commits to a decision on $15$--$30\%$ of borderline content ($n_\text{kept} = 150$--$303$ out of $1{,}000$); the rest are routed to human review at a $7$--$25\times$ higher volume than OCSVM-v4. Disabling the abstain mechanism (forcing the classifier to commit on every input) does not resolve the trade-off: at the natural decision threshold, MLP-v4's borderline FPR jumps to $26$--$47\%$, an order of magnitude above OCSVM-v4 in two of three domains. \textbf{There is no operating point where MLP-v4 simultaneously commits to the majority of borderline traffic and maintains FPR comparable to OCSVM-v4.}

\paragraph{The bimodal score distribution is the structural cause.} The MLP head's near bimodal output (probabilities clustered near $0$ or $1$ on confident records, with a thin uncertain band in between) is what drives both failure modes. The ambiguous middle band coincides with the borderline-safe content the detector is asked about: records in this band trigger the abstain rule (high abstain volume), and when forced to commit, those same records cross $P(\text{unsafe}){>}0.5$ from below and manifest as the high full-population FPR. The trade-off is structural: any abstain rule based on classifier confidence produces an analogous coverage/FPR exchange.

\paragraph{Threshold-free accuracy is comparable, not the deciding axis.} On AUROC the three architectures sit within $\Delta \leq 0.07$ ($0.92$--$1.00$, computed on the kept population). The MLP achieves the highest AUROC ($\sim\!0.99$) but, as the previous paragraph established, only realizes that accuracy on the small subset it commits to. The LLM-judge matches OCSVM-v4 on AUROC ($0.92$--$0.94$ vs.\ $0.93$--$0.97$), ruling out the interpretation that a $14$B prompted model cannot discriminate QI clusters; its limitation is calibratability and unit cost, not discrimination.

\paragraph{LLM-judge fails on calibration and unit cost.} The LLM-judge's discrete output distribution prevents tuning to a target operating point: the achievable operating points are restricted to the handful of values the prompted model emits. Combined with $\sim\!1$\,s latency and $\sim\!\$10^{-2}$ per-query cost ($200\times$ slower, $10^{4}\times$ more expensive than the trained detectors), this disqualifies the LLM-judge from inline Layer-2 deployment regardless of accuracy.

\paragraph{Why the one-class detector wins?} OCSVM-v4 is the only configuration in Table~\ref{tab:baselines} that simultaneously (a)~commits to a decision on $\geq 89\%$ of borderline content, (b)~remains conservative at the natural threshold whether abstain is on or off ($\leq 36\%$ borderline FPR in the worst domain, both rows), (c)~exposes a calibrated abstain region without a coverage cliff, and (d)~runs at millisecond latency. The supervised classifier wins on AUROC but only by routing the majority of borderline traffic to review; the LLM-judge wins on no operational axis. For the review-queue-bounded, sub-second-latency Layer-2 regime characterized in Section~\ref{sec:overview}, the one-class detector is the deployment-correct architecture.

\section{Discussion}
\label{sec:disc}

The three-domain evaluation (Section~\ref{sec:eval}) and the three-architecture comparison (Section~\ref{sec:comparison}) together support three claims: the case-style confound is a property of the synthetic training regime rather than of any single domain or estimator; the data-plus-method fix that closes it is portable across domains; and the deployment choice between architectures is dominated by the abstain-rate/FPR trade-off, not by AUROC. We unpack each below and close with the operational consequences for production Layer-2 deployment.

\subsection{What the Three-Domain Evaluation Establishes}
\label{sec:disc:generalization}
Three findings transfer cleanly across medical, finance, and law:

\paragraph{The diagnostic transfers.} The within-distribution-vs-borderline AUROC gap exceeds $0.15$ in every domain under T3+GMM-v3 (Table~\ref{tab:borderline}, Figure~\ref{fig:results}), with the borderline AUROC band $[0.72, 0.79]$ entirely disjoint from each domain's within-distribution AUROC ($\geq0.95$ in medical and finance, $0.95$ in law). Three independently instantiated corpora, three QI taxonomies, three borderline-safe constructions: one signature. We treat this as strong evidence that the borderline-safe stress test should be a default evaluation ingredient for any one-class density detector trained on synthetic data.

\paragraph{The combined-fix recipe transfers.} T3+OCSVM-v4 lifts borderline AUROC to $\geq 0.93$ in all three domains and shifts the entire ROC tail ($\Delta \text{FPR}_{95} = 44$--$55$\,pp, $\Delta \text{FPR}_{90} = 43$-$56$\,pp). That both operating points improve by similar margins rules out target-TPR-specific artifacts. The recipe (T3+OCSVM estimator + Mistral-generated borderline-aug) is portable.

\paragraph{Cross-generator robustness transfers within the open $7$-$14$B class.} A held-out 9B-class generator from a distinct lineage (Yi-1.5-9B-Chat) is detected at AUROC $\geq 0.95$ in all three domains under T3+OCSVM-v4 (Table~\ref{tab:crossgen}). For medical and finance, cross-generator AUROC matches or exceeds within-distribution; for law, the slight gap ($-0.013$) sits at the natural threshold, not the threshold-free discriminator (Appendix~\ref{app:strata}). We make no claim about closed-source frontier generators or gradient-guided adversarial generation.

\paragraph{What does \emph{not} transfer: the per-subtype residual.} Medical and law converge to a $\sim\!4\%$ mean borderline FPR after the fix; finance retains $\sim\!35\%$ (Table~\ref{tab:borderline:strata}). Law shares finance's structural register-dominance property (a single conversational seed at $\geq 74\%$ of the safe corpus), yet exhibits the medical-shape residual. Part of finance's elevated $\tau=0$ residual is operating-point artifact (finance's $\tau=0$ sits at 97.8\% TPR vs medical's 80.2\%, Section~\ref{sec:eval:borderline}); the gap that survives at matched TPR is the genuine cohort-quantitative-content overlap. The three-domain comparison therefore \emph{rules out} register dominance as a sufficient cause and localizes the finance residual to legitimate cohort-scoped quantitative content (income bands, contribution limits) that is stylistically indistinguishable from single-client QI signals in the embedding space. A two-domain study could not have made this distinction visible. Mitigating finance's residual requires a domain-specific augmentation or a quantitative-content gating filter (Section~\ref{sec:limitations:finance:residual}).

\subsection{Why the One-Class Architecture Wins?}
\label{sec:disc:architecture}
The cross-architecture comparison (Section~\ref{sec:comparison}, Table~\ref{tab:baselines}) reframes what ``best detector'' means at deployment scale. The supervised binary classifier (MLP-v4) is more accurate on AUROC ($\sim\!1.00$ vs.\ $\sim\!0.953$-$0.97$), and the LLM-judge sits withing $\sim0.05$ AuROC of T3+OCSVM-v4, yet \emph{neither} is deployable in the regime we target. The supervised classifier's near-bimodal score distribution forces it into a binary trap: either it abstains on $69$-$85\%$ of borderline content, or it flags $26$-$47\%$ of safe content at the natural threshold. The LLM-judge produces only a handful of discrete output values, preventing operating-point calibration, and runs $200\times$ slower at $10^{4}\times$ the per-query cost.

T3+OCSVM-v4 is the only architecture that simultaneously (a)~commits to a decision on $\geq 89\%$ of borderline records, (b)~remains conservative at the natural threshold ($\leq 36\%$ borderline FPR even in the worst domain), (c)~exposes a calibrated abstain region, and (d)~runs at single-digit-millisecond latency. The architectural property that earns the win is the T3+OCSVM's graduated signed-distance score: unlike T3+GMM log-likelihoods that saturate on tail content, and unlike binary cross-entropy probabilities that compress most of the borderline mass into the $0.4$-$0.6$ band, the T3+OCSVM produces a continuous score that remains numerically meaningful on inputs the training distribution did not anticipate. This is the same property that makes T3+OCSVM-v4 dominate T3+GMM-v3 on the borderline stress test within our own architecture (Section~\ref{sec:eval:ablation_analysis}).

\subsection{Operating-Point Selection and the Abstain Gate}
\label{sec:disc:operating}

The T3+OCSVM-v4 detector exposes two operating regimes without retraining (Section~\ref{sec:detection:discrim}): conservative ($\tau{=}0$) and balanced (FPR@$95\%$TPR). Their numerical profile is per-domain.

\begin{itemize}
    \item \textbf{Medical / law:} the conservative point is the right default. TPR is $72$-$80\%$, within-distribution FPR is $\leq 0.7\%$, borderline FPR is $\leq 3.1\%$. Suitable for clinician- and attorney-facing tooling where over-flagging causes alert fatigue.
    \item \textbf{Finance:} the conservative point is \emph{not} recommended where borderline content is common (borderline FPR $36.0\%$); use the balanced point with downstream review.
\end{itemize}

The abstain region of Section~\ref{sec:detection:abstain} is the deployment-time response to inputs both densities consider out-of-distribution. Under T3+GMM-v3, medical and finance borderline abstain rates sit at $\sim\!30\%$, which is operationally infeasible at scale (Table~\ref{tab:abstain}). T3+OCSVM-v4 reduces these rates $3$-$8\times$ in medical and finance; law's T3+GMM-v3 rate is already low ($4.6\%$, see Section~\ref{sec:eval:abstain}) and changes only marginally. Abstain volume is therefore not a footnote but a \emph{second} deployment-time metric on which T3+OCSVM-v4 dominates T3+GMM-v3, complementing the binary FPR reduction. Operating-point selection is therefore a per-domain engineering decision against measured borderline profile and downstream cost asymmetry; we recommend reporting both points in deployment documentation and revisiting them periodically.

\subsection{Practical Takeaways}
\label{sec:disc:takeaways}
For practitioners building Layer-2 detectors over RAG outputs:

\begin{enumerate}
    \item \textbf{Evaluate against borderline-safe content.} Within-distribution AUROC is misleading for synthetically-trained one-class detectors; the borderline-safe stress test should be a default evaluation ingredient.
    \item \textbf{Prefer T3+OCSVM over T3+GMM under stylistic overlap}, and choose a one-class density architecture over a supervised binary classifier when the deployment budget for human review is bounded.
    \item \textbf{Augment the safe side with borderline-safe records.} A modest $4{,}000$-record borderline-aug corpus (generated by a different model than the one used at evaluation) lifts borderline AUROC by $0.14$-$0.24$ in our three domains.
    \item \textbf{Calibrate operating points per-domain}, and add an explicit abstain gate for inputs that are out-of-distribution to both densities.
\end{enumerate}

\section{Limitations and Future Work}
\label{sec:limitations}

We highlight four limitations of the current work and close with additional future-work directions.

\subsection{Domain and Language Scope}\label{sec:limitations:domain}
The evaluation covers English-language medical, finance, and legal Q\&A content. The detector's behavior on other clinical-data modalities (radiology, discharge summaries, structured EHR), other regulatory regimes (GDPR's broader notion of personal data, civil-law jurisdictions, non-US finance regimes), and other languages is unmeasured. Each new domain requires its own QI taxonomy, validator extras, and seed corpus, and Section~\ref{sec:disc:generalization} shows that per-subtype residual structure can be domain-dependent (Section~\ref{sec:limitations:finance:residual}). The present three-domain evidence supports the claim that the \emph{methodology} (axis-stratified synthetic generation, dual-detector OOD, borderline-safe stress test) transfers, but not that detector-tuning is universal.

\subsection{Finance Borderline Residual}\label{sec:limitations:finance:residual}
Finance T3+OCSVM-v4 borderline FPR at $\tau{=}0$ is $36.0\%$, versus $2.4\%$ medical and $3.1\%$ law. The three-domain comparison (Section~\ref{sec:disc:generalization}) localizes the cause to legitimate quantitative content overlap on three borderline subtypes: cohort-scoped financial statements (income bands, contribution limits, dollar amounts) are stylistically indistinguishable from single-client QI signals. Two corrective directions:
\begin{itemize}
    \item \textbf{Quantitative-content gating.} A pre-filter that strips numeric ranges and currency vocabulary before scoring would partially address the overlap, at the cost of careful design to avoid stripping legitimate QI content.
    \item \textbf{Discriminative classifier (Section~\ref{sec:limitations:discriminative}).} A fine-tuned encoder model trained directly on the safe/unsafe label would optimize for QI-presence discrimination rather than density estimation, which we expect would push finance borderline performance closer to medical/law levels.
\end{itemize}

Two ancillary finance-specific notes: (i) the $80\%$ Money StackExchange seed dominance contributes to the wider safe distribution, but does not by itself explain the residual; (ii) the $4{,}000$-record borderline-aug scale matches medical and law but may be insufficient for finance's wider stylistic surface, thus needs an augmentation-size ablation study in the future.

\subsection{Synthetic Training Data Biases}\label{sec:limitations:synthetic}
The unsafe class is constructed entirely through LLM generation against curated specifications. Three biases follow: (i)~\textbf{generator-style residuals} (mitigated within the open 7--14B class by the cross-generator result of Section~\ref{sec:eval:crossgen}, but unmeasured against closed-source frontier models); (ii)~\textbf{plausibility ceiling} (generators occasionally produce domain-implausible answers; we did not measure clinical/financial/legal validity of generated unsafe records); (iii)~\textbf{adversarial-mode imbalance} ($\textsc{indirect\_qi}$ rejection rates of $\sim$80\% mean its share in the training corpus is below the uniform-axis-sampling target). Each bias has a tractable mitigation (plausibility filter, adversarial oversampling, per-mode retry budget tuning) but is not closed in the current work.

\subsection{One-Class Density Ceiling and Discriminative Alternative}\label{sec:limitations:discriminative}
The dual one-class density approach achieves AUROC $\geq 0.96$ on within-distribution and held-out-generator data in all three domains but the borderline AUROC remains in $[0.93, 0.97]$ even with the combined fix, with a residual $36\%$ borderline FPR at the natural threshold in the finance domain. We hypothesize this reflects a fundamental ceiling of the paradigm: density estimators do not directly optimize for class separation, and the QI signal is intrinsically lower-dimensional than the stylistic signal that shapes the safe/unsafe densities.

The frozen-encoder discriminative alternative is already characterized in this work as MLP-v4 (Section~\ref{sec:comparison}); its near-bimodal output forces an unacceptable coverage/FPR tradeoff at deployment scale. The remaining unexplored alternative is a fully fine-tuned encoder model (DistilBERT-class) trained end-to-end on the safe/unsafe label, where the representation itself is optimized for QI-presence discrimination. We expect this to narrow the gap at the borderline, particularly in finance, but it sacrifices the encoder-agnostic, label-free deployment property that motivated the one-class formulation.

\subsection{Other Future-Work Directions}\label{sec:limitations:future}
Three additional directions are worth highlighting: \textbf{verbatim-source leakage} (a second sub-type of contextual privacy leakage where RAG outputs near-verbatim quotes retrieved from documents, not addressed by the current generator); \textbf{multi-turn leakage} (QI clusters revealed across turns rather than within a single answer; requires conversation-level scoping); and \textbf{adversarial robustness} (gradient-guided attacks against the embedding pipeline or density estimators are not characterized, and the $\textsc{indirect\_qi}$ adversarial mode probes only static paraphrasing). Per-replication caveats also remain: law's within-distribution AUROC ($0.9711$, Table~\ref{tab:crossgen}) is mildly optimistic because no separate within-distribution unsafe holdout was reserved from the Yi slice, the $131$-name \texttt{COMMON\_NAMES} list is empirically grown rather than exhaustive, and FinanceBench's $90$-record size limits per-source stratified analysis.

\section{Conclusion}
\label{sec:conclusion}

We presented a Layer~2 contextual privacy detector for RAG systems
based on a dual one-class density estimator over fused text
embeddings, and evaluated it across three domains (medical, finance,
law) that span four jurisdictions and three stylistic registers.
Three findings are load-bearing.

First, within-distribution AUROC on synthetic-vs-synthetic Q\&A is
misleading: across three independently constructed domains the
T3+GMM-baseline detector achieves $\geq 0.95$ AUROC within-distribution
but collapses to $[0.72, 0.79]$ AUROC on a borderline-safe stress
test, exposing that it has learned referential client/patient/matter
voice as a proxy for quasi-identifier presence. The borderline-safe stress test we introduce to expose this confound is a methodological contribution that extends beyond contextual privacy to any classifier trained on synthetically constructed positive and negative classes.

Second, a T3+OCSVM estimator on the safe side with a $4{,}000$-record borderline-safe augmentation in training recovers borderline AUROC to $\geq 0.93$ in all three domains and reduces borderline FPR$_{95}$ by $44$--$55$\,pp across all three
domains, with comparable FPR$_{90}$ reductions. The recipe is portable; per-subtype residual structure is
domain-dependent.

Third, the third domain (law) was decisive in localizing the cause of the elevated finance residual: law shares finance's structural
register-dominance property but exhibits the medical-shape residual,
ruling out register dominance as a sufficient explanation and
identifying finance-specific quantitative content overlap as the
attributable cause. The three-domain study makes this distinction
visible; a two-domain study could not.

The framework is not closed: a discriminative classifier should
narrow the residual further, particularly in finance; multi-turn
leakage and verbatim-source leakage are unmodeled, and gradient-guided
adversarial users with white-box access to the detector are
uncharacterized. Each is a tractable direction for future work
(Section~\ref{sec:limitations}). What the present three-domain evidence supports is the claim that the methodology and the diagnostic generalize, the recipe is portable, and the residual gaps are
localizable to specific structural properties of the safe distribution
in each new domain.

\clearpage

\bibliographystyle{IEEEtran}
\bibliography{refs} 

\clearpage

\appendix

\section{Per-Domain Detail}
\label{app:domains}

This appendix collects per-domain texture that did not fit in the
merged main sections: per-domain motivation and corpus rationale
(Section~\ref{app:motivation}), full per-generator generation tables
(Section~\ref{app:gen}), full ablation grid for medical
(Section~\ref{app:ablation}), per-generator stratified TPR for the T3+OCSVM-v4
configuration in each domain (Section~\ref{app:strata}), and per-domain
borderline-safe corpus construction notes (Section~\ref{app:borderline}).

\subsection{Domain-Selection Rationale}
\label{app:motivation}

\paragraph{Why finance was selected as the second domain.}
Among candidate non-medical domains we considered finance, law,
education, and HR records. Finance was selected on five criteria:
(i)~quasi-identifier crispness (regex/checksum-validatable direct
identifiers like SSN, ABA routing, EIN, Luhn-validatable card
numbers); (ii)~validator tractability (finance Q\&A typically uses
``Client A'' placeholders rather than fictional named individuals,
unlike legal exam hypotheticals); (iii)~public-corpus availability
(Money StackExchange, FinanceBench, FinQA all under open licenses);
(iv)~commercial threat model (retail banking chatbots, robo-advisors,
fintech support agents are large RAG deployment surfaces under GLBA,
FCRA, PCI~DSS); (v)~hypothesized confound (finance has stylistic
analogs to medical's case voice: ``this client'', ``the household
we're reviewing'', ``in this scenario'').

\paragraph{Why law was selected as the third domain.}
Law was deferred from the second-domain decision on validator-design
grounds: legal exam hypotheticals routinely use fictional first-name
parties (``Alice sues Bob'') that conflict with the shared
name-based PII detector. With finance established as a reproducing
second domain, we revisited law and addressed the validator obstacle
directly: (i)~the shared \texttt{COMMON\_NAMES} list was expanded
from 53 to 131 entries to cover crypto-tutorial conventions and
gap-filler first names; (ii)~surname-only case citations
(\textit{Brown v.\ Board}) were preserved as legitimate legal
vocabulary by requiring either two title-cased tokens after a
recognized first name or an explicit title prefix; (iii)~three
law-specific direct-identifier extras (federal/state docket numbers,
keyword-anchored bar numbers, keyword-anchored attorney license
numbers) were added. With these changes, the validator passes
$\textit{Brown v.\ Board of Education}$ but rejects ``a recent widow,
Jane'' as a name leak.

A two-domain study could have stopped at ``finance is harder''. The
third domain made one specific structural cause of the finance
residual visible: register dominance is shared by law and finance,
but only finance carries the elevated residual.

\subsection{Per-Generator Generation Campaigns}
\label{app:gen}

Each domain's training-pool generation is split across three
generators in a uniform $15{,}000 + 1{,}500 + 3{,}000 = 19{,}500$
records per generator pattern: a $15{,}000$-record baseline (no
adversarial mode), $1{,}500$ records of \textsc{indirect\_qi}, and
$3{,}000$ records of mixed
\textsc{distractor\_padded}+\textsc{style\_transfer}. The held-out
generator (Yi-1.5-9B) contributes $2{,}000$ baseline records per
domain. The borderline-aug generator (Mistral, used \emph{only} on
the safe side) contributes $4{,}000$ records per domain across the
five borderline-safe subtypes; the borderline-eval generator (Phi-3,
held out from training) contributes $1{,}000$ records per domain.

\paragraph{Validator yields.}
Baseline yield is $\geq 95\%$ in all three domains. \textsc{indirect\_qi}
yield is $20$--$40\%$ depending on domain (lowest in medical, highest
in law), reflecting domain-specific difficulty in producing
QI-paraphrased answers that survive the forbidden-vocabulary check.
\textsc{distractor\_padded} and \textsc{style\_transfer} yields are

\paragraph{Per-axis distribution after validation.}
Within each domain's post-validator unsafe corpus, all six framings,
all four placements, and all eight QI types are represented within
$\pm 5\%$ of the configured uniform-sampling targets. The QI-count
distribution $(2, 3, 4) = (0.45, 0.35, 0.20)$ is realized as
approximately $(0.53, 0.34, 0.13)$, with the slight skew toward
two-QI records reflecting higher rejection rates at higher QI counts
in the \textsc{indirect\_qi} mode.

\subsection{Full Ablation Grid (Medical)}
\label{app:ablation}

The merged Section~\ref{sec:eval} reports the v3 vs.\ v4 endpoints of the
$2 \times 2$ ablation grid (estimator $\in \{$T3+GMM, T3+OCSVM$\}$, safe
data $\in \{$v3, v4$\}$). The full medical grid is reproduced here.

\begin{table}[h]
\centering
\caption{\textbf{Full medical ablation grid.} Each row is one detector
configuration (estimator $\times$ safe-side training data; v3 = seed
corpus only, v4 = seed corpus + 4k borderline-aug). Each column is one
test pairing: within-distribution, cross-generator (Yi held out),
borderline-safe stress test. Bold rows are the v3 baseline and v4
combined fix reported in the main text. The intermediate
configurations (T3+GMM-v4, T3+OCSVM-v3) are reported here for medical only;
finance and law evaluate only the v3 and v4 endpoints
(Section~\ref{sec:setup:configs}). Reported on the full population
to keep all four cells comparable; the bold v3/v4 endpoints shift
by $\leq 0.01$ AUROC under the kept-only convention used by
Tables~\ref{tab:withindist}, \ref{tab:crossgen}.}
\label{tab:app:ablation}
\footnotesize
\setlength{\tabcolsep}{4pt}
\begin{tabular}{llccc}
\toprule
\textbf{Estimator} & \textbf{Safe} & \textbf{Within-dist} & \textbf{Cross-gen} & \textbf{Borderline} \\
\midrule
\textbf{T3+GMM}   & \textbf{v3} & \textbf{0.9930} & \textbf{0.9940} & \textbf{0.7179} \\
T3+GMM            & v4          & 0.9874          & 0.9879          & 0.9364          \\
T3+OCSVM          & v3          & 0.9958          & 0.9943          & 0.8976          \\
\textbf{T3+OCSVM} & \textbf{v4} & \textbf{0.9922} & \textbf{0.9943} & \textbf{0.9575} \\
\bottomrule
\end{tabular}
\end{table}

The ablation isolates each fix's contribution. On the borderline
stress test (the binding metric), data alone (T3+GMM-v4) lifts AUROC
from $0.7179$ to $0.9364$
($+0.218$); estimator alone (T3+OCSVM-v3) lifts it to $0.8976$
($+0.180$); the combination (T3+OCSVM-v4) lifts it to $0.9575$
($+0.240$). The two fixes are not strictly additive but are mutually
reinforcing.

We did not retrain the equivalent $4$-cell grid for finance and law
because the medical grid already characterizes each fix's
contribution and the cross-domain replication targets the v3-vs-v4
endpoints. A finance-specific ablation that would test whether
finance shows a different ratio of method-level vs.\ data-level
contribution is left to future work
(Section~\ref{sec:limitations:finance:residual}).

\subsection{Per-Generator Stratified TPR (T3+OCSVM-v4)}
\label{app:strata}

\begin{table}[h]
\centering
\caption{Per-generator TPR at $\tau{=}0$ under T3+OCSVM-v4 across the
three domains. Yi-1.5-9B is held out from unsafe-side training in all
three domains, so its row reports clean cross-generator behavior.
The training-pool spread is small in all three domains
($\leq 5$\,pp). For medical and finance, Yi is detected at
within-pool-comparable TPR; for law, Yi sits noticeably below the
training-pool mean, even though its threshold-free AUROC is still
high (Table~\ref{tab:crossgen}).}
\label{tab:app:strata}
\small
\begin{tabular}{lccc}
\toprule
\textbf{Generator} & \textbf{Medical} & \textbf{Finance} & \textbf{Law} \\
\midrule
Qwen2.5-7B-Instruct      & 0.796 & 0.971 & 0.720 \\
Mistral-7B-Instruct-v0.3 & 0.773 & 0.974 & 0.696 \\
Phi-3-medium-4k-instruct & 0.798 & 0.968 & 0.742 \\
\midrule
Yi-1.5-9B-Chat \emph{(held out)}      & 0.801 & 0.992 & 0.511 \\
\midrule
\textbf{Spread (training pool)} & 2.5\,pp & 0.6\,pp & 4.6\,pp \\
\textbf{Yi vs.\ training mean}  & $+0.012$ & $+0.021$ & $-0.208$ \\
\bottomrule
\end{tabular}
\end{table}

The narrow training-pool spread rules out single-generator
stylometric overfitting in each domain. For medical and finance, the
held-out generator is detected at \emph{higher} TPR than the training
mean, consistent with Yi-1.5-9B producing more stereotypical
referential-voice phrasing on the same prompts than the training
pool. For law the pattern inverts: Yi sits $\sim$21\,pp below the
training-pool mean at $\tau{=}0$. Because the threshold-free
discriminator is still strong (cross-gen AUROC $0.9580$,
Table~\ref{tab:crossgen}), this represents a shift in the operating
point at which Yi crosses the safe/unsafe boundary,  not a failure
of the detector to separate Yi outputs from safe content. We
attribute the shift to law's wider safe distribution and Yi's
stylistically distinct legal register relative to the training pool,
and to the fact that medical and finance training pools may have
overfit slightly to \emph{any} stereotypical 7--14B output style
(making Yi unusually easy), where law's generators produce more
diverse legal phrasing in the training pool, leaving less room for Yi
to look more stereotypical.

\subsection{Borderline-Safe Construction Notes}
\label{app:borderline}

The five borderline-safe subtypes are instantiated per domain with
explicit prompt templates that specify (i)~the QI-free constraint,
(ii)~the stylistic register to imitate, and (iii)~five canonical
example openings. Each borderline-aug record is sampled by a
training-pool model (Mistral) at temperature $0.9$ and validated
against the same direct-identifier prohibitions as the unsafe corpus
plus the additional ``no QI cluster of size $\geq 2$'' rule.
Borderline-eval records are sampled by a held-out model (Phi-3) under
the same constraints. The $4{,}000$/$1{,}000$ split in scale
reflects the v4 augmentation requirement (training-aug needs to be
large enough to shift the T3+OCSVM safe boundary) versus the
borderline-eval requirement (eval needs to be large enough to
estimate FPR with reasonable variance).

\paragraph{Per-domain borderline subtype counts.}
Across all three domains the $1{,}000$-record borderline-eval is
approximately balanced across the five subtypes
($\sim 200$ records each, varying $\pm 12$ records due to
generator-side rejection patterns). Per-domain n's are reported in
Table~\ref{tab:borderline:strata} via the n column of the underlying
per-subtype tables.

\end{document}